%% file: arxiv.tex
\documentclass[twoside,11pt]{article}

\usepackage[preprint]{jmlr2e}

\usepackage{microtype}
\usepackage{graphicx}
\usepackage{subfigure}
\usepackage{booktabs}

\usepackage[utf8]{inputenc} 
\usepackage{amsmath}
\usepackage{amssymb}
\usepackage{bbm}
\usepackage{xcolor}
\usepackage{tabularx}
\usepackage{hyperref}
\usepackage{paralist}
\usepackage{multirow}
\usepackage[ruled]{algorithm2e}
\usepackage{algorithmic}
\input{macros}

\DeclareMathOperator*{\argmin}{arg\,min}
\DeclareMathOperator*{\argmax}{arg\,max}

\newcommand{\qedsymbol}{\ensuremath{\blacksquare}}

\jmlrheading{1}{2022}{1-48}{4/00}{10/00}{}{Matthias Feurer, Benjamin Letham, Frank Hutter and Eytan Bakshy}

\ShortHeadings{Practical Transfer Learning for Bayesian Optimization}{Feurer, Letham, Hutter and Bakshy}
\firstpageno{1}

\begin{document}

\title{Practical Transfer Learning for Bayesian Optimization}

\author{\name Matthias Feurer \email feurerm@cs.uni-freiburg.de \\
    \addr Department of Computer Science, Faculty of Engineering\\
    University of Freiburg, 79110 Freiburg, Germany
    \AND
    \name Benjamin Letham \email bletham@fb.com \\
    \addr Meta\\
    1 Hacker Way, Menlo Park, CA 94025, USA 
    \AND
    \name Frank Hutter \email fh@cs.uni-freiburg.de \\
    \addr Department of Computer Science, Faculty of Engineering\\
    University of Freiburg, 79110 Freiburg, Germany, and \\
    Bosch Center for Artificial Intelligence, Germany
    \AND
    \name Eytan Bakshy \email ebakshy@fb.com \\ 
    \addr Meta \\
    1 Hacker Way, Menlo Park, CA 94025, USA 
}

\editor{}

\maketitle

\begin{abstract}
When hyperparameter optimization of a machine learning algorithm is repeated for multiple datasets it is possible to transfer knowledge to an optimization run on a new dataset. We develop a new hyperparameter-free ensemble model for Bayesian optimization that is a generalization of two existing transfer learning extensions to Bayesian optimization and establish a worst-case bound compared to vanilla Bayesian optimization. Using a large collection of hyperparameter optimization benchmark problems, we demonstrate that our contributions substantially reduce optimization time compared to standard Gaussian process-based Bayesian optimization and improve over the current state-of-the-art for transfer hyperparameter optimization.
\end{abstract}


\section{Introduction}
Bayesian optimization (BO) is a data-efficient blackbox optimization technique that is routinely used for hyperparameter optimization (HPO) of machine learning (ML) algorithms~\citep{snoek-nips12a,feurer-bookchapter19a}. Given a small initial set of function evaluations, BO proceeds by fitting a surrogate model to those observations, typically a Gaussian process (GP), and then optimizing an acquisition function that balances exploration and exploitation in determining what point to evaluate next.

If the same ML algorithm with the same hyperparameters is routinely optimized for different datasets, for example in online machine learning services, we can lift the blackbox assumption and use data from prior BO runs as ancillary information. As the hyperparameter surfaces of an ML algorithm applied to different datasets are similar, we aim for a strategy to transfer this ancillary information 
across datasets
to find good solutions faster than 
vanilla BO.
To facilitate such transfer learning in BO we specify several desiderata to allow broad applicability to new applications:

\begin{enumerate}
    \item{Scalability to many tasks:} having many potentially related past BO runs requires a method that is able to handle both many tasks and many observations in total across all tasks.
    \item{Fast update of the surrogate model:} similarly, such a method needs to be able to easily update the surrogate model after adding a new observation on the current target task. 
    \item{Easy and fast adaptability to different tasks:} the method should neither depend on task-specific hyperparameter settings, nor depend on numerical task descriptors which have to be manually defined by experts.
    \item{No performance degradation:} transfer learning must not substantially worsen performance compared to vanilla BO, even in the worst case of misleading prior information.
\end{enumerate}

Even though several BO methods have been developed to borrow strength across runs (see Section \ref{sec:pastwork} for a review), none of them satisfy all of the desiderata above.
The main merit of this work is to introduce a set of methods that does so.
Our specific contributions are:\footnote{An earlier version of this work is available on arXiv and we compare against it in Appendix~\ref{app:older}.}

\begin{enumerate}
    \item We develop a new ensemble model for BO, based on a linear combination of GPs and show that this is a generalization of two existing, scalable transfer learning extensions to BO and is similarly scalable (desiderata \#1 and \#2).
    \item We introduce a novel, hyperparameter-free (desideratum \#3) method to weight models in a transfer learning setting based on Agnostic Bayesian Learning of Ensembles, which allows us to bound the worst-case slowdown over BO to a multiplicative factor (desideratum \#4).
    \item We perform the largest experimental evaluation of transfer learning for hyperparameter optimization (transfer HPO) to date, using six hyperparameter optimization benchmarks with a total of 249 tasks and \numcompetitors{} competitor methods from the literature. Our contributions robustly improve over the current state of the art, without any need for hyperparameter tuning or benchmark-specific modifications.
    \item We provide the implementation of our methods, the reimplementations of the competitor methods, and the collection of benchmarks as open-source code, to both simplify further research and reproducing our experiments.
\end{enumerate}

The paper is structured as follows. First, in Section~\ref{sec:background} we give background information on BO, introduce the problem setting and give a primer on related work. Then, in Section~\ref{sec:linear_combination} we review related work on linear combinations of models for Bayesian optimization. Next, we present our methodology in Section~\ref{sec:methodology}. Then, we describe the baselines and benchmarks we release (Section~\ref{sec:code}) and provide experimental evidence that our proposed methodology is preferable (Section~\ref{sec:experiments}). Finally, we give a comprehensive overview of related work (Section~\ref{app:related}) before concluding the paper (Section~\ref{sec:conclusion}).

\section{Background and Problem Setup}\label{sec:background}

In this section we provide the necessary background and introduce the problem setup for the related work on linear combinations of models in Section~\ref{sec:linear_combination} and our own, new methodology in Section~\ref{sec:methodology}.

\subsection{Bayesian Optimization}

The goal of BO is to find a global minimum $\mathbf{x}_{opt}$ of a blackbox function in a bounded space $\mathcal{X}$ by iteratively querying the function at input configurations $\mathbf{x}_1, \mathbf{x}_2, \dots, \mathbf{x}_n$ and observing the corresponding outputs $y_1, y_2, \dots, y_n$. In each iteration, we first fit a probabilistic model $f$ on observations $\mathcal{D} = \{(\mathbf{x}_k, y_k)\}_{k=1}^n$ made so far. We then use an acquisition function $\alpha(\mathbf{x})$ to select a promising configuration to evaluate next, balancing exploration and exploitation.

In general, $y_k$ may be a noisy estimate of the function value. 
We estimate the underlying function with GP regression, yielding a posterior $f(\mathbf{x} | \mathcal{D})$ with an analytical mean $\mu(\mathbf{x})$ and variance $\sigma^2(\mathbf{x})$~\citep[Eqs. 2.25 and 2.26]{rasmussen-book06a}:
\begin{equation}\label{eq:GP}
    \mu(\mathbf{x}_*) = \mathbf{k}_*^T(\mathbf{K} + \sigma^2_n I)^{-1}\mathbf{y}, \hspace{1cm}
    \sigma^2(\mathbf{x}_*) = k_\mathbf{\theta}(\mathbf{x}_*,\mathbf{x}_*) - \mathbf{k}_*^T(\mathbf{K} + \sigma^2_n I)^{-1}\mathbf{k}_*,
\end{equation}

\noindent{}where $k_\mathbf{\theta}(\cdot,\cdot)$ is a kernel function with hyperparameters $\theta$, $\mathbf{k}_*$ is the vector of covariances between the new point $\mathbf{x}_*$ and all training points, and $\mathbf{K}$ is the kernel matrix of size $n \times n$ where each entry $\mathbf{K}_{i,j} = k_\mathbf{\theta}(\mathbf{x}_i, \mathbf{x}_j)$.

The most commonly used acquisition function is the expected improvement (EI)~\citep{jones-jgo98a}, which can be computed in closed form and generally gives good results. Let $f(\mathbf{x}_{\textrm{best}})$ be the current best function value: $f(\mathbf{x}_{\textrm{best}}) = \min_{k\in(1, \ldots, n)} f(\mathbf{x}_k)$.
The EI is
\begin{equation}
\label{equ:ei}
    \alpha(\mathbf{x}  | \mathcal{D}) = \mathbb{E}_{f(\mathbf{x} | \mathcal{D})} \left[\max(0, f(\mathbf{x}_{\textrm{best}}) - f(\mathbf{x})) \right] 
     = \sigma(\mathbf{x}) z \Phi(z) + \sigma(\mathbf{x}) \phi(z),
\end{equation}
with $z = \frac{f(\mathbf{x}_{\textrm{best}}) - \mu(\mathbf{x})}{\sigma(\mathbf{x})}$, $\Phi(\cdot)$ being the CDF and $\phi(\cdot)$ being the PDF of the normal distribution.
Thorough introductions to Bayesian optimization are given by \citet{brochu-arXiv10a}, \citet{shahriari-ieee16a}, \citet{frazier-arxiv2018a} and \citet{garnett-bayesoptbook22a}.

\subsection{Problem Setup}

We suppose that $t-1$ BO runs have been completed on previous tasks with the same search space $\mathcal{X}$. Let $\mathcal{D}^i = \left\{(\mathbf{x}^i_k, y^i_k)\right\}_{k=1}^{n_i}$ be the function evaluations made for past optimization run $i \in (1,\dots,t-1)$. We fit a GP model to the observations of each past run $i$ and refer to these models as \textit{base models}. They have posterior $f^i(\mathbf{x} | \mathcal{D}^i)$, with mean and variance $\mu_i(\mathbf{x})$ and $\sigma^2_i(\mathbf{x})$, respectively. These remain fixed throughout the optimization, inasmuch as we do not obtain new observations for them. 
The current optimization problem we are trying to solve is task $t$. We fit a GP to observations from task $t$ and call it the \textit{target model} $f^t$. The target model is refit after each new function evaluation. We overload notation and define $\mathcal{D} = \{\mathcal{D}^1, \ldots, \mathcal{D}^t\}$. Our goal is to minimize the target function using the base models $f^1, \dots, f^{t-1}$ and the target model $f^t$.

\subsection{A Primer on Related Work}\label{sec:pastwork}
Borrowing strength from past runs is a form of meta-learning and transfer learning. Techniques for transfer learning for hyperparameter optimization (transfer HPO) can be categorized into 9 approaches: 1) a single model that is trained on all tasks simultaneously, 2) a learned kernel that is trained on the base tasks and applied to the target task, 3) learning an adaptation of each base task to the current task, 4) an initial design learned from previous optimization runs, 5) reducing the design space based on previous optimization runs, 6) changing the acquisition function to include past data or models, 7) learning a separate model for each base task and combining those, 8) hybrid methods, and 9) prior-based methods. We review these techniques in Section~\ref{app:related} and refer to \citet{vanschoren-automl2019a} for an in-depth overview of meta-learning. We discuss two relevant methods which fall into categories 6 and 7 in Section \ref{sec:linear_combination}, since they are special cases of the methods we introduce in this paper.

\section{Weighting-Based Bayesian Optimization Transfer Learning}\label{sec:linear_combination}

Several recent works in transfer learning for BO suggest modeling the current task $t$ as a linear combination of a surrogate model fit to current task $t$, as well as one surrogate model fit to each of the base tasks $(1,\dots,t-1)$ \citep{wistuba-ecml16a,lindauer-aaai18,wistuba-ml18a}. If the fitted base models were stored during past runs, they can be used directly without any refitting of prior results, and only the GP for the current task needs to be updated. 
As an additional advantage, such methods do not require the existence of meta-features. However, they depend crucially  on the strategy for computing the weights.
In this section, we review the ranking method employed by the two-stage transfer surrogate for learning this linear combination and an acquisition function that is used on top of such an ensemble, and assess them with respect to the design criteria given in the introduction.

\subsection{Two-Stage Transfer Surrogate Model with Rankings}\label{sec:linear_combination_of_models}

The two-stage transfer surrogate model with rankings \citep[\tstr{},][]{wistuba-ecml16a} suggests combining the mean predictions of the models as 
\begin{equation*}
    \bar{\mu}(\mathbf{x}_*) = \sum_{i=1}^t w_i \mu_i(\mathbf{x}_*),
\end{equation*}
while dropping the variance of the base models, and using for a predictive variance just that of the target model.
It uses a Nadaraya-Watson kernel weighting to linearly combine the predictions from each GP according to a normalized distance measure across tasks: 
\begin{equation*}
    w_i = \frac{k_{\rho}(\mathbf{\chi}_i,\mathbf{\chi}_t)}{\sum_{j=1}^t k_{\rho}(\mathbf{\chi}_j,\mathbf{\chi}_t)},    
\end{equation*}
with the kernel
\begin{equation*}
    k_{\rho}(\mathbf{\chi}_i,\mathbf{\chi}_j) = \gamma \left( \frac{||\mathbf{\chi}_i - \mathbf{\chi}_j||_2}{\rho} \right).
\end{equation*}
Here $\rho$ is a bandwidth hyperparameter and $\gamma$ is given by the Epanechnikov kernel: 
\begin{equation*}
    \gamma(r) = 
    \begin{cases}
        \frac{3}{4} (1 - r^2) & \text{if } r \leq 1 \\
        0 & \text{otherwise}.
    \end{cases}
\end{equation*}

The most important question is how to describe each task with a vector $\mathbf{\chi}$, and \citet{wistuba-ecml16a} proposed two different methods to obtain them: (1)~descriptive meta-features~\citep{brazdil-ecml94a,rivolli-arxiv19a}, and (2)~directly modeling the distance as the proportion of discordant pairs when the base model is evaluated on configurations evaluated on the current task:
\begin{equation*}
|| \mathbf{\chi}_i - \mathbf{\chi}_t||_2 = \frac{2}{n_t(n_t - 1)}\sum_{k = 1}^{n_t - 1} \sum_{l = k + 1}^{n_t} \mathbbm{1}((f^i(\mathbf{x}_k^t) < f^i(\mathbf{x}_l^t)) \oplus (y_k^t < y_l^t)),
\end{equation*}
where $\oplus$ is the exclusive-or operator.

The bandwidth $\rho$, a sensitive hyperparameter which must be chosen by the user (violating desideratum \#3), gives the fraction of discordant pairs a previously observed dataset may share with the target dataset before being discarded.
Updating the distance measure over time improves over the meta-feature based distance measure (as also shown by \citet{leite-mldmpr12} for a closely related method), and we will therefore not compare to any meta-feature-based method (requiring meta-features would also violate desideratum \#3).

The kernel is used to combine mean predictions of base models with the mean prediction of the target model, but variances are not combined---the combined model is given the variance of the target model and variances of base models are ignored.
Furthermore, \tstr{} assigns a constant value to the distance of the target model to itself; therefore the actual weight of the target model in the ensemble depends on the magnitude of the weights of the base models.
As weights are computed independently of each other, correlated base tasks can shift the weight in the combined model.
Due to the close relation to our proposed model, we include \tstr{} for comparison in our experiments of Section \ref{sec:experiments}, where we will refer to using the model as a drop-in replacement of the standard GP in BO as \tstrei{}.

\subsection{Transfer Acquisition Function}\label{sec:taf}

Another approach to handling meta-data in BO is to adapt the acquisition function to a so-called \emph{transfer acquisition function} \citep[TAF,][]{wistuba-ml18a}. Similarly to \tstrei{}, the TAF uses a weighting of the different tasks, but applies them to the acquisition function rather than to the surrogate model. The expected improvement acquisition function using only the target model $f^t$ on the target task is combined linearly with an improvement term computed on each base task:

\begin{align*}
    \alpha(\mathbf{x}) &= w_t EI_t(\mathbf{x}) + \sum_{i=1}^{t-1} w_i I_i(\mathbf{x}) \\
    &= w_t EI_t(\mathbf{x}) + \sum_{i=1}^{t-1} w_i \max \left(0, \left(\min_{\mathbf{x}_k \in \mathcal{D}^t}f^i(\mathbf{x}_k)\right) - f^i(\mathbf{x}) \right) \\
    &= w_t \mathbb{E}_{f^t(\mathbf{x})} \left[\max(0, f^t(\mathbf{x}_{\textrm{best}}) - f^t(\mathbf{x})) \right] + \sum_{i=1}^{t-1} w_i \max \left(0, \left(\min_{\mathbf{x}_k \in \mathcal{D}^t}f^i(\mathbf{x}_k)\right) - f^i(\mathbf{x}) \right),
\end{align*}

\noindent{}where $f(\mathbf{x}_{\textrm{best}}) = \min_{k\in(1, \ldots, n^t)} f^t(\mathbf{x}_k^t)$, i.e. the best observation on the target task as predicted by the target model.

The idea is to fade out base tasks more quickly than \tstrei{} by not only measuring the similarity with the target task as discussed in the previous subsection, but also by measuring whether there is still information that can be transferred (i.e., whether the base model ``knows" about a better hyperparameter setting than those tried so far).

The TAF uses the \tstr{} weighting scheme which we described in the previous subsection, and inherits the bandwidth hyperparameter $\rho$. As TAF is a special case of the model we develop in this paper, we will include it as a baseline in Section \ref{sec:experiments}. We discuss this method in more detail in Appendix~\ref{app:active_testing}, where we also relate it to the well-known meta-learning strategy of \emph{active testing}~\citep{leite-mldmpr12}.

\section{Methodology}\label{sec:methodology}

In this section, we describe our new approach for transfer learning in BO. We first develop a new probabilistic weighting scheme for combining base models with the target model (Section \ref{sec:weighting}). Next, we demonstrate how it can be used in a linear combination of GPs and present how to use this ensemble model for BO (Section~\ref{sec:rgpe}). Then, we introduce a regularization mechanism to deal with a potentially large number of unrelated base models (Section \ref{sec:dilution}). We end this section by providing theoretical insight into the described weighting scheme and regularization (Section \ref{sec:theory}).

\subsection{Computing Ensemble Weights}\label{sec:weighting}

Intuitively, we wish to learn a stacking meta-model that combines all base models and the target model into a single, joint model~\citep{pardoe-icml10a}. 
Training data for the stacking meta-model is generated in two ways.
For base models, we evaluate their ability to predict the observations $\mathcal{D}^t$ for task $t$.
For the target model, these would be in-sample predictions and thus overly optimistic, so we use leave-one-out cross-validation on $\mathcal{D}^t$ to get an estimate of the target model's ability to correctly predict the target task.
Then, a combined model is learned to explain the target task observations in the best possible way.
In the BO setting, the target model has access to data from the correct function (the target task), but it will not be the best model to use early in the optimization because it does not yet have enough data to accurately model the function.
Therefore, for each timestep, we need to find out which models perform best at explaining the data observed so far.
In the beginning, base models trained on previous tasks might perform better than the target model, so we would like to use them to transfer knowledge. As soon as we are certain that the target model is the best, we would like to rely on it instead.

A solution for this problem is given by agnostic Bayesian ensemble learning~\citep[\emph{ABE},][]{lacoste-icml14a}. 
Given a desired loss function, each predictor in the ensemble is weighted according to the probability that it is the best predictor in the ensemble for the target task.\footnote{ABE was introduced to construct an ensemble from models trained on a single dataset where we have access to a validation dataset or can obtain it via cross-validation. In contrast, we use ABE here to transfer knowledge from prior tasks to new tasks in a transfer stacking manner and have therefore adjusted the way we obtain this validation set.}
This has the desirable property of taking into account our uncertainty about when to trust the target model and which base models to rely on. This strategy further allows us to also learn the weight of the target model, which is not possible with the \tstr{} strategy.
Concretely, ABE constructs an ensemble as
\begin{equation*}
    \bar{f}(\mathbf{x}) = \sum_{i=1}^t p(f^i == f^*|\mathcal{D}^t) f^i(\mathbf{x}),
\end{equation*}
where $f^* \in \{f^1, f^2, \dots, f^t\}$ is the predictor that minimizes the desired loss function.

ABE either directly models the posterior risk of a potential ensemble member or uses the bootstrap of the predictions on the target tasks; both of these approaches are agnostic to the models used.
We use the latter method due to its simplicity and because it only requires the models' predictions on the target task. 

We first construct the loss function for use with ABE. Our goal in BO is to find the minimum function value and a model will be useful for optimization if it is able to correctly order observations according to their function value. For meta-learning, we wish to assess the ability of model $i$ to generalize to the target function, and so construct a loss function that measures the degree to which each model is able to correctly rank the target observations $\mathcal{D}^t$. Given $n_t > 1$ target function evaluations, we define the loss as the number of misranked pairs:
\begin{equation}\label{eq:loss}
\mathcal{L}(f, \mathcal{D}^t) = \sum_{k = 1}^{n_t} \sum_{l = 1}^{n_t} \mathbbm{1}((f(\mathbf{x}_k^t) < f(\mathbf{x}_l^t)) \oplus (y_k^t < y_l^t)),
\end{equation}
\noindent{}where $\oplus$ denotes the exclusive OR operator (XOR).

Note here that we are evaluating base models only on their predictions for the target task, and thus are assessing their ability to generalize to the target function. For the target model, as mentioned above, this would be an estimate of the in-sample error and would not accurately reflect generalization. We estimate generalization in the target model using cross-validation, in practice with leave-one-out models. Let $f^t_{-j}$ indicate the target model with observation $(x^t_j,y^t_j)$ left out. The loss for the target model is computed as
\begin{equation}\label{eq:loss_target}
    \mathcal{L}(f^t, \mathcal{D}^t) = \sum_{k = 1}^{n_t} \sum_{l = 1}^{n_t} \mathbbm{1} ( (f^t_{-k}(\mathbf{x}_k^t) < y^t_l )
   \oplus (y_k^t < y_l^t)).
\end{equation}

Models $f^t_{-j}$ are only useful when we fit them with at least two observations. For this reason we start the weighting procedure only when we have gathered three observations. Before that we assign uniform weights to all models.

The ranking loss is preferable to other choices, such as squared error or model log-likelihood, because the actual values of the predictions do not matter for optimization---we only need to identify the location of an optimum. It is easy to see that if all of the models are able to correctly order a set of points then the ensemble will also correctly order those points. 

We now weight each model with the probability that it is the ensemble member with the lowest ranking loss. 
We estimate this probability with a Monte Carlo approximation. Namely, we draw $S$ bootstrap samples $\ell_{i,s} \sim \mathcal{L}(f^i, \mathcal{D}_{s}^{t})$ from each of our models $i=1, \ldots, t$, and for each sample $s$ assess which model aligns best with the observed data $\mathcal{D}^t$.
The weight for model $i$ is then computed as

\begin{equation} \label{eq:weight}
    w_i = \frac{1}{S}\sum_{s=1}^S \left( \frac{\mathbb{I}(i \in \argmin_{i'}l_{i',s})}{\sum_{j=1}^t \mathbb{I} (j \in \argmin_{i'}l_{i',s})} \right),
\end{equation}

\noindent{}which distributes the weight of a sample across all members of the argmin in case of a tie. Because base models compete against each other, correlated base models are weighted down as in the original ABE strategy~\citep{lacoste-icml14a}. Using ABE induces a slight overhead compared to \tstr{}, but is still substantially faster than using a single Gaussian process for all data points (see Appendix~\ref{sec:methodology:scaling}). We present pseudo-code in Algorithm~\ref{alg:weights}.

\begin{algorithm}[tbp]
\begin{small}
\caption{Learning weights with the agnostic Bayesian learning of ensembles}
  \label{alg:weights}
\begin{algorithmic}[1]
    \STATE {\bfseries Input:} Number of bootstrap samples $S$, all observations on all base tasks and the target task $\mathcal{D}^i \forall i \in (1,\dots,t)$, all base models and the target model $f^i \forall i \in (1,\dots,t)$
    \IF{$|\mathcal{D}^t| < 3$}
        \STATE \textbf{return} weights $\left[\frac{1}{t} | i \in (1,\dots,t) \right]$
    \ENDIF
    \STATE $L = [[l_{1,1}, \dots, l_{1,S}],\dots,[l_{t,1},l_{t,S}]]$ \tcp*[h]{Initialize $2d$ array to store the losses per model and bootstrap sample}
    
    \FOR{$s \in (1, 2, \dots, S)$}
        \STATE $\mathcal{B} = \{randint(1, n_t)\}_{k=1}^{n_t}$ \tcp*[h]{Draw bootstrap indices with replacement}
        \STATE $\mathcal{D}^t_s = \left\{(\mathbf{x}^t_k, y^t_k)|k \in \mathcal{B}\right\}$ \tcp*[h]{Resample the target task data}
        \FOR{$i \in (1, 2, \dots, t)$}
        
            \IF{$i < t$}
                \STATE $l_{i,s} = \mathcal{L}(f^i,\mathcal{D}^t_s)$ \tcp*[h]{Equation~\ref{eq:loss}}
            \ELSE
                \STATE $l_{t,s} = \mathcal{L}(f^t,\mathcal{D}^t_s)$ \tcp*[h]{Equation~\ref{eq:loss_target}}
            \ENDIF
        
        \ENDFOR
    
    \ENDFOR
    \STATE Compute weights $\mathbf{w} = \left[ \frac{1}{S}\sum_{s=1}^S \left( \frac{\mathbb{I}(i \in \argmin_{i'}l_{i',s})}{\sum_{j=1}^t \mathbb{I} (j \in \argmin_{i'}l_{i',s})} \right) | i \in (1,\dots,t) \right]$ \tcp*[h]{ Equation~\ref{eq:weight}}
    \STATE \textbf{return} weights $\mathbf{w}$
\end{algorithmic}
\end{small}
\end{algorithm}

\subsection{Ranking-Weighted Gaussian Process Ensemble}\label{sec:rgpe}

Having established a weighting strategy, we propose to use a weighted combination of the predictions of each base model and the target model itself: 
$
    \bar{f}(\mathbf{x} | \mathcal{D}) = \sum\nolimits_{i=1}^t w_i f^i(\mathbf{x} | \mathcal{D}^i).
$
We will show that this is a very powerful formalism. First, this ensemble model remains a GP, and in particular
\begin{equation}\label{eq:sum}
    \bar{f}(\mathbf{x} | \mathcal{D}) \sim \mathcal{N} \left(\sum\nolimits_{i=1}^t w_i \mu_i(\mathbf{x}), \sum\nolimits_{i=1}^t w_i^2 \sigma_i^2(\mathbf{x}) \right),
\end{equation}
which leaves the usual computational machinery for BO with GPs valid, such as the applicability of arbitrary acquisition functions, a closed-form expression for EI, and the ability to draw joint samples for parallel optimization. We dub this model ranking-weighted Gaussian process ensemble, or short, \rgpe{}.

Additionally, each base model remains unchanged throughout the optimization and can be loaded directly from the previous runs. The fitting cost is only the cost of fitting the target model and inferring the weights $w_i$.
Finally, predictions are made independently for each GP, and at prediction time we obtain only a linear slowdown relative to using only the target model. 
In case there would be too many data points for the base tasks to fit a GP on them, one could also replace them with more scalable models, such as Bayesian neural networks~\citep{schilling-ecmlpkdd15a,snoek-icml15a,springenberg-nips16a} or random forests~\citep{hutter-lion11a}, but this is beyond the scope of this paper. 

To compensate for the fact that response values of different tasks can live on different scales, we standardize each model prior to inclusion in the ensemble~\citep{yogatama-aistats14a}. We  study alternatives to this strategy in Appendix~\ref{ssec:copula}.

We now present how we will use the \rgpe{} model for BO and give an overview of the different instantiations of \rgpe{} in Table~\ref{tab:methods_rgpe}.

\begin{table}[]
    \centering
\begin{tabular}{lp{0.65\textwidth}p{0.1\textwidth}}
\toprule
Abbreviation & Description & Section \\
\midrule
\rgpemean{} & Ranking-weighted Gaussian process ensemble using only the variance of the target model & \ref{sec:ei} \\
\rgpenei{} & Ranking-weighted Gaussian process ensemble with noisy EI & \ref{sec:nei} \\
\rmogp{} & Mixture of Gaussian processes weighted by \rgpe{} & \ref{sec:rmogp} \\
\tafrgpe{} & Transfer acquisition function with the \rgpe{} model & \ref{sec:tafrgpe} \\
\bottomrule
\end{tabular}
    \caption{Instantiations of \rgpe{}.\label{tab:methods_rgpe}}
\end{table}

\subsubsection{Expected improvement and Dropping Base Model Variance}\label{sec:ei}

The most straight-forward application of \rgpe{} for BO is to use the ensemble as a plug-in replacement for the standard GP. As the most simple strategy one can use regular expected improvement with the posterior mean as a plug-in estimate for $f(\mathbf{x}_{\textrm{best}})$. As a second ad-hoc strategy we consider using only the variance of the target model and drop the base model variances as suggested by \tstrei{} (see Section~\ref{sec:linear_combination}).\footnote{
An alternative derivation of dropping the base model variances is using the weighted base models as a mean function in the GP model. If we define the mean function to be $m(\mathbf{x}) = \sum_{i=1}^{t-1} w_i f^i(\mathbf{x})$ and the target GP to regress $\tilde{y} = m(\mathbf{x}) + w_t y$ we obtain a predictive mean of the form

\begin{equation}
    \mu(\mathbf{x}_*) = m(\mathbf{x}_*) + \mathbf{k}_*^T(\mathbf{K} + \sigma^2_n I)^{-1}(\mathbf{\tilde{y}} - \mathbf{m}(\mathbf{X})), \\
\end{equation}
while the variances of the base models are not considered at all.} We refer to these techniques as \rgpeei{} and \rgpemean{}.

\subsubsection{Noisy Expected Improvement}\label{sec:nei}

However, with the formulation in Equation \ref{eq:sum}, the base models can add uncertainty to points where the target model is certain. In particular, they can add posterior variance to points where the target model has noiseless observations, which can inflate the acquisition value around already-observed points. Because base models are not updated over time, this could lead to the optimization process stalling when using standard EI as we discussed in the previous subsection. Also, if observations are noisy or if there is uncertainty in base models at the current best, $f(\mathbf{x}_{\textrm{best}})$ in the expected improvement (see Equation~\ref{equ:ei}) is not a constant, but rather is a random variable with uncertainty both in which point is the best, and what the function value at that point is. This can occur when the locations of the observations in the base models do not overlap with those of the target model or if there is overlap but noise in the base models, or if there is noise in the target model. 

Typical approaches for computing the expected improvement with noisy observations can be used with the \rgpe{}; we use the so-called noisy expected improvement to account for the uncertainty in the best value~\citep{letham-ba19a}. We refer to the resulting combination of \rgpe{} and the noisy expected improvement as \rgpenei{}.

With this approach, we integrate out the uncertainty in the current best value, extending the expectation in Equation~\ref{equ:ei} to be over the joint posterior of $p(f(\mathbf{x}), f(\mathbf{x}_{\textrm{best}}) | \mathcal{D})$. We first describe the procedure for a single Gaussian process before explaining how to use it with the \rgpe{} model.

The expectation is computed by first using Monte-Carlo (MC) sampling to integrate over the posterior $p(f(\mathbf{x}_{\textrm{best}}) | \mathcal{D})$, by drawing joint samples from the ensemble posterior at all of the observations $\mathbf{x}_1,\mathbf{x}_2,\dots,\mathbf{x}_{n}$. In each sample, $f(\mathbf{x}_{\textrm{best}})$ is deterministically computed as the best value in the sample draw. For each joint sample, we can then compute the conditional posterior $p(f(\mathbf{x}) | f(\mathbf{x}_{\textrm{best}}), \mathcal{D})$ by conditioning the Gaussian process on the (noiseless) sample values at the observations. This means that we condition the GP on $\{(\mathbf{x}_k,\tilde{f}(\mathbf{x}_k)\}_{k=1}^{n}$ with $\tilde{f} \sim p(f|\mathcal{D})$.\footnote{In practice, for numerical stability the noise levels at these conditioning points will not be exactly 0, so we set it to the minimum noise level our GP can take on during its hyperparameter optimization: $e^{-25}$.} Afterwards, $f(\mathbf{x}_{\textrm{best}})$ is deterministic and we can compute EI for each MC sample in the usual manner using Equation~\ref{equ:ei}. Averaging EI over the MC samples produces an estimate for the joint expectation. 

For use with an ensemble model the noisy EI requires us to  sample from the ensemble posterior. Because all models in the ensembles we consider are independent, the sampling and conditioning described above can be done independently for each base model. The integration procedure does not change for the target model, but for the base models. Here we need to draw joint samples at the locations observed on the target task and the respective base tasks, i.e. conditioning the GP for task $i$ on $\{(\mathbf{x}_k^i,\tilde{f}^i(\mathbf{x}_k^i)\}_{k=1}^{n^i} \cup \{(\mathbf{x}_k^t,\tilde{f}^i(\mathbf{x}_k^t)\}_{k=1}^{n^t}$ with $\tilde{f}^i \sim p(f^i|\mathcal{D}^i)$. For each draw we then computed EI using the ensemble. In practice we used $30$ draws of quasi-Monte Carlo integration (using scrambled Sobol sequences as described by \cite{letham-ba19a}). 

\subsubsection{Ranking-Weighted Mixture of Gaussian Processes}\label{sec:rmogp}

Having introduced a combination of models based on the addition of random variables, we now establish a way in which the combination of GPs can be used in a mixture-of-GPs fashion~\citep{tresp-neurips01a}.
The PDF of a Gaussian mixture model at $f(\mathbf{x})$ with $t$ components is $p_{\textrm{mix}}(f(\mathbf{x})) = \sum_{i=1}^t w_i p_i(f(\mathbf{x}))$, where $p_i$ is the (Gaussian) PDF of component $i$. Making use of the fact that we can sample from a mixture-of-GPs by first sampling a mixture component and then sampling from the GP, the expected improvement at $\mathbf{x}$ can be decomposed:
\begin{equation}
    EI_{\textrm{mix}}(\mathbf{x}) = \mathbb{E}_{f(\mathbf{x}) \sim p_{\textrm{mix}}}[I(\mathbf{x})]
    = \mathbb{E}_{i} \mathbb{E}_{f(\mathbf{x}) \sim p_{i}}[I(\mathbf{x})] = \sum\nolimits_{i=1}^t w_i EI_{i}(\mathbf{x}),
    \label{eq:rmogp}
\end{equation}
where $EI_{i}(\mathbf{x})$ is the EI computed under mixture component $i$. Thus, EI for the mixture model is a weighted sum of the EI for each of member of the ensemble. In order to select $\mathbf{x}_{\textrm{best}}$ for $EI_i$ we use the predictive mean of the respective model $i$. Importantly, we choose $\mathbf{x}_{\textrm{best}} \in \mathcal{D}^t$, so we aim to improve only over observations made on the target task.
We call the resulting procedure \rmogp{}.

In order to deal with noise in the base models we use the re-interpolation trick~\citep{forrester-book2008}. 
The re-interpolation trick can be regarded as a one-shot alternative to the noisy EI described in Section~\ref{sec:nei}. Instead of using MC integration, we condition the base model GP $f^i$ on the predictive mean at the target observations, i.e. $\mathcal{D}^i \cup  \{(\mathbf{x}_k^t,f^i(\mathbf{x}_k^t)\}_{k=1}^{n_t}$. This ensures that the individual $f^i(\mathbf{x}_{best})$ per base model are constants and the EIs using the individual base models can be computed in closed form.

Additionally, the result in Equation \ref{eq:rmogp} makes it clear how the mixture of GPs can be used with any other acquisition function that is based on the expectation of a quantity~\citep{wilson-neurips18a}, for example the knowledge gradient~\citep{frazier-arxiv2018a} and max-value entropy search~\citep{wang-icml17a}, while it is not clear whether this can be done with the TAF.

\subsubsection{Transfer Acquisition Function}\label{sec:tafrgpe}

Finally, we notice a striking resemblance between this method and the transfer acquisition function from Section \ref{sec:taf}, with a single difference: Equation \ref{eq:rmogp} uses the EI and not only the improvement for the base models. The transfer acquisition function is therefore a different approach of dealing with the variance of the base models in a mixture-of-GPs, and this line of reasoning gives us a derivation of the TAF from our \rgpe{} model given in Equation~\ref{eq:sum}. This combination of our \rgpe{} model with the well-performing transfer acquisition function will be the overall best method in our experiments and we will refer to it as \tafrgpe{}. 

\subsection{Preventing Weight Dilution}\label{sec:dilution}

One challenge with this type of ensemble is preventing weight dilution by a large number of models. Due to the bootstrap sampling, even models with bad generalization performance have a chance of obtaining a non-zero weight. 
This is because when we consider a very large number of poor models, the chance of at least one of them producing the correct ranking on a sample goes to $1$ as the number of models increases.

In order to prevent weight dilution we add a base model filtering step to drop base models which are unlikely to improve over the target model.
The weight calculation mechanism presented in the previous section allows us to directly compare two models according to their quality, and in particular to compare the quality of any base model to the quality of the current target model.
We prevent weight dilution by discarding base models with frequency proportional to their probability of producing a worse ranking than the target model. Base model $i \in (1,\dots,t-1)$ is discarded from the ensemble in each iteration with probability 
\begin{equation}\label{equ:p_drop}
    p_{drop}(i) = 1 - \left( \left(1 - \frac{n_t}{H} \right) \frac{\sum_{s=1}^S \mathbbm{1}(l_{i,s} < l_{t,s})}{S} \right),
\end{equation}
where $H$ is the optimization horizon. The factor $\frac{\sum_{s=1}^S \mathbbm{1}(l_{i,s} < l_{t,s})}{S}$ is the probability that the base model outperforms the target model. 
Instead of a conjugate beta prior, we use a multiplicative prior $(1 - \frac{n_t}{H})$ that reduces the probability of keeping a base model linearly over time and encodes our belief that the target model is the correct model and should be favored. This ensures that at the end of the optimization we use only the target model. We highlight that this mechanism does not automatically give a higher weight to the target model. Base models are discarded prior to computing the weights. Therefore, the target model still needs to outperform the remaining base models in order to obtain a high weight. 

Preventing weight dilution also has computational benefits in that it results in fewer GP predictions for each ensemble prediction.  
Models are only removed from the ensemble proportionally to their probability of performing worse than the target model. In contrast, \tstr{} removes models if they perform badly on the observed function evaluations without taking into account the performance of the target model, rather based on the hyperparameter $\rho$. Furthermore, there is a non-zero probability that we only consider the target model in each iteration, which we will use in Section~\ref{sec:theory}.

\subsection{Theoretical Analysis}\label{sec:theory}

Here, we show that in each iteration the proposed weighting mechanism has a positive chance of performing vanilla BO, in which only the target model is used. 
Therefore, standard proofs for the convergence of BO apply~\citep{bull-jmlr11a} with a multiplicative slowdown factor.\footnote{This only applies to results that give guarantees in terms of the simple regret and not the cumulative regret.}

\begin{theorem}
    Bayesian optimization using a linear combination of Gaussian processes with weights learned according to Section~\ref{sec:dilution} is at most a factor of 
    \begin{equation*}
        1 / \left( \frac{1}{H} \sum_{h=1}^{H} \left(\frac{h}{H}\right)^{t-1} \right)    
    \end{equation*}
    slower than Bayesian optimization in the worst case.
\end{theorem}

As before, $H$ is the optimization horizon, while we change the number of observed data points $n^t$ to $h$, i.e. the current iteration.

\emph{Proof sketch:} In order for the proposed method to fall back to vanilla BO, we need the weights of all models except that of the base model to be zero. Given the definition of $p_{drop}$ in Equation \ref{equ:p_drop}, and setting $n_t$ to $h$ in $p_{drop}(i,h)$, we can calculate the probability of 
dropping all base models at step $h$ as $\prod_{i=0}^{t-1} p_{drop}(i,h)$,
and so the expected proportion of iterations that proceed as vanilla BO is

\begin{align}\label{equ:bound1}
    & \frac{1}{H} \sum_{h=1}^{H} \prod_{i=1}^{t-1} p_{drop}(i,h)\\
    & = \frac{1}{H} \sum_{h=1}^{H} \prod_{i=1}^{t-1} \left(1 - \left(\left(1 - \frac{h}{H}\right) \frac{\sum_{s=1}^S \mathbbm{1}(l_{i,s}  < l_{t,s})}{S}\right)\right)\\
    & \ge \frac{1}{H} \sum_{h=1}^{H} \left(1 - \left(\left(1 - \frac{h}{H}\right) \frac{S}{S}\right)\right)^{t-1} \\
    & = \frac{1}{H} \sum_{h=1}^{H} \left(\frac{h}{H}\right)^{t-1}\\
    & > 0
\end{align}
The observations gathered in iterations when not all base models are dropped do not impose any issues on the convergence proof by Bull, which we show in Appendix~\ref{app:proof}.~~~~~~~~~~~~~~~~~\qedsymbol{}

This analysis is for the worst-case setting in which every base model consistently outperforms the target model. 
In practice, base models are frequently worse than the target model, leading to an improved bound.

\section{An Experimental Framework for Transfer Hyperparameter Optimization}\label{sec:code}

In this section we describe the experimental framework we implemented to develop our new methods and compare them to prior work from the literature. For this, we give an overview of the benchmark problems we use and also detail our implementation of the proposed methods and prior work into the Bayesian optimization framework SMAC3.

We provide all code at \url{https://github.com/automl/transfer-hpo-framework}. 

\subsection{Benchmarks}

We evaluate our methods on six benchmark problems we collected from the literature. We make them available with a uniform API to simplify their usage and allow them to be reused.

First, we use two HPO benchmark problems for optimizing the four hyperparameters of a support vector machine (\svmfd{}) and the ten hyperparameters of gradient boosting (\xgb{}) following \citet{perrone-neurips18a}. These benchmark problems are surrogate benchmarks built on data from \url{OpenML.org}~\citep{vanschoren-sigkdd14a}. They were not released to the public, but data to construct the surrogates is available from OpenML.org, and so we have reconstructed them and made them publicly available as part of our framework release.\footnote{We use the curated data by \cite{kuehn-arxiv18a}, which is actually a superset of what was used by \citet{perrone-neurips18a}. This data is available in multiple csv files created by the uploaders of the data to OpenML, while \cite{perrone-neurips18a} downloaded the data from OpenML.org independently. These csv files contain all results required to construct the surrogates.} These are evaluations on 38 binary datasets without missing values from the OpenML 100~\citep{bischl-arxiv17a}. We follow the standard methodology of employing a random forest as a surrogate~\citep{eggensperger-aaai15}, and optimize its hyperparameters using random search and 10-fold cross-validation to maximize the Spearman rank correlation. We then also use the data from the same source that is available for generalized linear models to construct a 2-d HPO problem (\glmnet{}). 

Second, we use LCBench, a neural network benchmark that contains the performance of deep neural networks implemented in Auto-Pytorch~\citep[\nn{}]{zimmer-tpami21a}. These networks are trained on 35 different tasks from the OpenML AutoML benchmark~\citep{gijsbers-automl19a}, excluding the four \emph{large} ones. For each dataset, the benchmark contains 2000 randomly sampled hyperparameter settings from a 7-d hyperparameter space.  

Third, we use two large sets of hyperparameter optimization benchmark problems for \adaboost{} and \svmsd{} on a diverse set of 50 datasets, with sizes ranging from 35 to 250000 training examples, and from 2 to 7000 features from~\citet{schilling-pkdd16a} and \citet{wistuba-ml18a}. Both of them are grid benchmarks, meaning that they contain a discretized grid of hyperparameter settings and the corresponding test-set accuracies. The \adaboost{} benchmark, inspired by a benchmark suggested by \cite{bardenet-icml13a}, contains a grid of two hyperparameters, the number of iterations and the number of product terms, for a total of 108 evaluations. The \svmsd{} benchmark contains a grid of six hyperparameters: three binary hyperparameters indicating a linear, polynomial, or RBF kernel; the penalty hyperparameter $C$; the degree of the polynomial kernel ($0$ if unused); and the RBF kernel bandwidth ($0$ if unused) for a total of 288 function evaluations. Note that this is a harder problem than the common 2-dimensional RBF SVM problem. 

All of these are available under a unified interface for the first time which allows them to easily be reused by other researchers. We summarize the benchmarks in Table~\ref{tab:benchmarks} and provide the exact search spaces in Appendix~\ref{app:experiments}.

\begin{table}[tb]
    \centering
    \small
    \label{tab:benchmarks}
    \begin{tabular}{lrrrrrrrr}
        \toprule
        Name & \rotatebox[origin=b]{60}{\# Dim} & \rotatebox[origin=b]{60}{\# Cat} & \rotatebox[origin=b]{60}{\# Cont} & \rotatebox[origin=b]{60}{\# Cond} & \rotatebox[origin=b]{60}{\# Tasks} & \rotatebox[origin=b]{60}{\# Samp} & \rotatebox[origin=b]{60}{Reference} \\
        \midrule
        \adaboost{} & 2 & 0 & 2 & 0 & 50 & 50 & \cite{schilling-pkdd16a} \\
        \svmsd{} & 6 & 3 & 3 & 0 & 50 & 50 & \cite{schilling-pkdd16a} \\
        \glmnet{} & 2 & 0 & 2 & 0 & 38 & 50 & ours \\
        \svmfd{} & 4 & 1 & 3 & 2 & 38 & 50 & \cite{perrone-neurips18a} \\
        \xgb{} & 10 & 1 & 9 & 4 & 38 & 50 & \cite{perrone-neurips18a} \\
        \nn{} & 7 & 0 & 7 & 0 & 35 & 50 & \cite{zimmer-tpami21a} \\
        \bottomrule
     \end{tabular}
         \caption{Summary of the benchmark problems considered. We give the name; dimensionality; number of categorical, continuous, and conditional hyperparameters; number of tasks; and the number of samples used to train the base models.
    }
\end{table}

\subsection{Implementation of Transfer HPO Methods}\label{sec:implemented-methods}

We used the SMAC3 Bayesian optimization package~\citep{lindauer-jmlr21a}, which to the best of our knowledge is the only BO package that natively supports Gaussian processes with categorical and conditional hyperparameters.
SMAC3 implements GPs with scikit-learn~\citep{scikit-learn}, and we used the ARD Mat\'{e}rn 5/2 kernel. 
Kernel hyperparameters $\theta$ were optimized in each iteration for the maximum a-posteriori estimate via L-BFGS-B using priors on the GP hyperparameters (a top-hat prior on the length-scales, a log-normal prior on the function scale and a horseshoe prior on the noise).
We used a Latin-hypercube initial design of size 10 for the GP-based vanilla BO.

For categorical hyperparameters, SMAC3 uses a Hamming kernel~\citep{Hutter09} and it uses a kernel separating the search spaces into subspaces if there are hyperparameters that are dependent on the exact setting of another hyperparameter~\citep{levesque-ieee17a}. In order to optimize the acquisition function for such mixed spaces, SMAC3 uses a stochastic local search starting from promising points in the search space~\citep{hutter-lion11a,levesque-ieee17a}. 
Obviously, this is not necessary for the grid-based benchmarks, for which we can simply compute the acquisition function value for each point. We used the expected improvement acquisition function~\citep{jones-jgo98a} in all experiments unless the method provides an alternate acquisition function itself (TAF, \rmogp{}) or requires special treatment of uncertainty (\rgpe{}). We give an overview over all implemented methods in Table~\ref{tab:methods}.

\subsection{Implemented Methods}\label{app:competitor-implementation}
We briefly describe the HPO methods we implemented. We start with single task \emph{standard HPO} baselines, then go on to prior \emph{transfer HPO} baselines from the literature, before finishing with the methods we propose in this paper. We give an overview of these methods in Table~\ref{tab:methods}. Following the design of SMAC3~\citep{lindauer-jmlr21a} we can implement them as four distinct building blocks of a Bayesian optimization algorithm: 
\begin{inparaenum}
    \item the hyperparameter search space,
    \item the initial design of Bayesian optimization,
    \item the model, and
    \item the acquisition function.
\end{inparaenum}

As standard HPO methods we provide a standard GP as described above and a Gaussian Copula process~\citep[GCP,][]{salinas-icml2020a}, in which the responses for each task are transformed to follow a standard normal distribution using a Copula transformation. Furthermore, we provide random search~\citep{bergstra-jmlr12a} and random search with simulated parallelism, which are suggested as sanity checks for hyperparameter optimization methods~\citep{recht-blog2016a} (Random 2x, 4x and 10x).\footnote{These are available as post-hoc methods. For each seed we perform random search once with 500 function evaluations and then postprocess each run to have 50 steps, where each step consists of running multiple evaluations (2, 4 or 10) in parallel in a synchronous fashion. We cannot run this for two of our grid-based benchmarks, \adaboost{} and \svmsd{}, as they have less than 500 points in the grid.}

We implemented the competitor methods search space learning~\citep{perrone-neurips19a}, sequential model-free hyperparameter optimization~\citep{wistuba-icdm2015a}, warmstarting algorithm configuration~\citep{lindauer-aaai18}, \ablr{}~\citep{perrone-neurips18a}, \gcpplusprior{}~\citep{salinas-icml2020a}, \tstrei{}~\citep{wistuba-ecml16a} and TAF~\citep{wistuba-ml18a} as closely as possible within our Python-based system. Methods not described in Section~\ref{sec:linear_combination_of_models} are described in Section~\ref{app:related}.

For the search space learning methods~\citep{perrone-neurips19a} there is no code available. We reimplemented the low-volume bounding box using for-loops (Section 4.2) and the handling of outliers using scipy~\citep{scipy} (Section 5). We then use it together with Random Search (\pruningwithrandomsearch{}) and a GP (\pruningwithBO{}).\footnote{Our implementation currently does not allow combining search space pruning with other transfer HPO methods.} Search space learning is implemented to run prior to calling Bayesian optimization and its only action is creating a new searchspace. Similarly, we had to reimplement sequential model-free hyperparameter optimization~\citep[\smfo{}]{wistuba-icdm2015a}, which then is an initial design passed to SMAC3.

For the \wac{} method~\citep{lindauer-aaai18} we used the publicly available code as a starting point for our implementation. 

\begin{table}[]
    \centering
\begin{tabular}{lp{0.4\textwidth}p{0.3\textwidth}}
\toprule
Abbreviation & Description & Reference \\
\midrule
GP & Standard Gaussian process & \cite{jones-jgo98a,lindauer-jmlr21a} \\
GCP & Gaussian process with Copula transform & \cite{salinas-icml2020a} \\
Random (1x) & Random search & \cite{bergstra-jmlr12a} \\
Random (2x,4x,10x) & Random search with simulated parallelism & \cite{recht-blog2016a} \\
\midrule
\pruningwithrandomsearch{} & Random search with a learned box-shaped search space & \cite{perrone-neurips19a} \\
\pruningwithBO & Gaussian process with a learned box-shaped search space & \cite{perrone-neurips19a} \\
\smfo{} & sequential model-free optimization & \cite{xu-aaai10a,xu-rcra11a,wistuba-icdm2015a,pfisterer-gecco21a} \\
\wac{} & warmstarting algorithm configuration & \cite{lindauer-aaai18} \\
\klweighting{} & Two-stage transfer surrogate model using KL-divergence based weighting & \cite{ramachandran-pkdd19a} \\
\ablr{} & adaptive basis function linear regression & \cite{perrone-neurips18a} \\
\gcpplusprior{} & Gaussian Copula process with a neural network prior & \cite{salinas-icml2020a} \\
\tstrei{} & Two-stage transfer surrogate model using pairwise rankings, see Section~\ref{sec:linear_combination_of_models} & \cite{wistuba-ecml16a} \\
\taftstr{} & Transfer acquisition function, see Section~\ref{sec:taf} & \cite{wistuba-ml18a} \\
\midrule
\rgpemean{} & Ranking-weighted Gaussian process ensemble using only the variance of the target model & ours, see Section~\ref{sec:ei} \\
\rgpenei{} & Ranking-weighted Gaussian process ensemble with noisy EI & ours, see Section~\ref{sec:nei} \\
\rmogp{} & Mixture of Gaussian processes weighted by \rgpe{} & ours, see Section~\ref{sec:rmogp} \\
\tafrgpe{} & Transfer acquisition function with the \rgpe{} model & ours, see Section~\ref{sec:tafrgpe} \\
\bottomrule
\end{tabular}
    \caption{Methods compared throughout this work. Top: standard HPO baselines. Middle: prior work on transfer HPO. Bottom: our methods.\label{tab:methods}}
\end{table}

For the linear weighting of models using the Kullback-Leibler divergence~\citep{ramachandran-pkdd19a} we used the MATLAB implementation provided by the authors as a starting point to reimplement the weighting scheme in Python.

For the \ablr{} method~\citep{perrone-neurips18a} there is no code available which we could either reuse or compare against and therefore used a Pytorch reimplementation.

For \gcpplusprior{} there is a reimplementation available by one of the authors\footnote{\url{https://github.com/geoalgo/A-Quantile-based-Approach-for-Hyperparameter-Transfer-Learning}}, but it was made available only after we wrote our own reimplementation using Pytorch. As the author's reimplementation does not achieve the exact numbers from the paper either, we stick to our own reimplementation.

While Java code is publicly available for \tstr{}~\citep{wistuba_code}, we obtained the code for TAF via private communication with the authors. 
\tstrei{} and TAF with the \tstr{} weighting (\taftstr{}) require setting a bandwidth hyperparameter $\rho$ and we choose it per benchmark to be the best on the remaining five benchmarks ($\rho \in (0.01, 0.02, 0.05, 0.1, 0.2, 0.3, 0.4, 0.5, 0.6, 0.7, 0.8, 0.9, 1.0)$). We refer to Appendix~\ref{ssec:implementation} for further details on our reimplementation.

Our code for these is organized as follows. For the approaches that learn an ensemble (\wac{}, \klweighting{}, \tstr{}) or are a single model (\ablr{}) we implement a new model class for SMAC3. The acquisition function is then implemented as a separate acquisition function class that can be used together with an ensemble. Because \gcpplusprior{} requires both a model and an acquisition function, we provide two classes that can be used together. All model-based methods use the same initial design and the minimal number of observations required to start the method from \smfo{} as a starting point. The minimal number is two for all methods that scale the observations to have zero mean and unit variance or using the copula transform and one for all methods that do not scale the data.

Finally, we implemented our methods as follows. We first implemented a model class for the \rgpe{} model described in Sections~\ref{sec:weighting} and~\ref{sec:dilution}, and second, implemented two acquisition functions, the noisy EI (\ref{sec:nei}) and the mixture of Gaussian processes (\ref{sec:rmogp}). This allows us to combine our ensemble with the noisy EI (\rgpenei{}), use our ensemble with regular EI (\rgpeei{}) and only the variance of the target model (\rgpemean{}), with the acquisition function based on the mixture of GPs (\rmogp{}) and also with the transfer acquisition function (\tafrgpe{}). For our methods we used $S=1000$ for each experiment and study its effect in Section~\ref{sec:ablation}.

\section{Experiments}\label{sec:experiments}

In this section we describe the experimental validation of the methods introduced so far. We first give the experimental setup and then we show and discuss the results. For better readability we split the presentation into three parts:
\begin{inparaenum}
    \item[1)] a comparison to prior work and baselines,
    \item[2)] a study on the evolution of weights over time, and
    \item[3)] an ablation study of our methods' components.
\end{inparaenum}
We provide further experiments on how to scale data, an ablation study on the effect of the hyperparameters of the baselines and comparisons to versions we proposed in earlier preprints of this paper in Appendix~\ref{app:experiments}.

\subsection{Experimental Setup}\label{sec:synthetic}

We optimize each function for a total of 50 iterations. 
For each meta-task we provide 50 function evaluations obtained by vanilla BO to the transfer HPO method. 
We then start each transfer HPO method with a single meta-learned configuration (or two if the method performs some sort of data scaling that requires at least two observations on the target task) and each other method with a Latin hypercube design of size 10. To account for randomness we conducted 15 repetitions.

We evaluated each benchmark in a leave-one-task-out fashion: we use one task as the target task and the remaining ones as base tasks.
For reporting comparable numbers, we first normalize the regret on each dataset between zero and one. Second, we average the normalized regret for all tasks within a benchmark. This metric is also known as the \emph{average distance to the global minimum}~(ADTM, \citep{wistuba-ml18a}). 

We provide results as tables and rescale the ADTM to be percentages (between zero and one hundred).
We typeset the best method in boldface and underline all methods which are not significantly different from the best according to a paired Wilcoxon signed-rank. 
For comparing performance, the \textit{paired} differences on the different tasks are the key quantity of interest, and the Wilcoxon signed-rank test provides a correct statistical test for assessing if one method performs better than another~\citep{demsar-06a}. The standard deviations in performance often used in BO papers are not useful here (and so are not given) because they include the significant variance from the different meta-datasets.

All experiments were conducted on a standard compute cluster with Broadwell E5-2630v4 2.2GHz processors.

\subsection{Comparison to Prior Work and Baselines}

In the first part of our experiments we compare the methods proposed in this paper to other popular hyperparameter optimization methods (standard HPO) and transfer learning hyperparameter optimization methods (transfer HPO) from the literature. We already described all methods in Section~\ref{sec:implemented-methods} and gave an overview in Table~\ref{tab:methods}. To fit all methods into a single table we have to split the results across three tables in total and present results for \glmnet{} and \svmfd{} in Table~\ref{tab:results-glm-svm}, results for \xgb{} and \nn{} in Table~\ref{tab:results-nn-xgb} and results for \adaboost{} and \svmsd{} in Table~\ref{tab:results-ada-svm}. For each method we give the averaged distance to the minimum at $10$, $20$, $30$, $40$ and $50$ iterations of BO.

On a high level, we find that our proposed methods (bottom groups) consistently improve over the standard HPO baselines (top groups) and transfer HPO baselines (middle groups).
Remarkably, \tafrgpe{} is either the best model or performs statistically equivalent to the best model on all six benchmarks. \tafrgpe{} is also able to outperform Random(10x) on three out of four benchmarks and is always better than Random(4x) (we cannot apply Random(2x) etc. on \adaboost{} and \svmsd{} as there are not sufficient points in the grid).

\begin{table}[tb]
    \centering
    \small
\begin{tabular}{lllllllllll}
\toprule
{} & \multicolumn{1}{c}{10} &  \multicolumn{1}{c}{20} & \multicolumn{1}{c}{30} & \multicolumn{1}{c}{40} & \multicolumn{1}{c}{50} & \multicolumn{1}{c}{10} &  \multicolumn{1}{c}{20} & \multicolumn{1}{c}{30} & \multicolumn{1}{c}{40} & \multicolumn{1}{c}{50} \\
\midrule
{} & \multicolumn{5}{c}{\glmnet{}} & \multicolumn{5}{c}{\svmfd{}} \\
\cmidrule(lr){2-6}
\cmidrule(lr){7-11}
GP     &                       1.92 &              \textbf{0.59} &              \textbf{0.45} &              \textbf{0.38} &              \textbf{0.35} &                       5.38 &                       1.60 &                       1.06 &                       0.76 &                       0.68 \\
GCP         &                       1.92 &              \textbf{0.55} &              \textbf{0.44} &              \textbf{0.39} &              \textbf{0.36} &                       5.38 &                       2.25 &                       1.78 &                       1.51 &                       1.33 \\
Random(1x)  &                       4.80 &                       1.40 &                       0.89 &                       0.70 &                       0.52 &                       5.60 &                       2.99 &                       2.12 &                       1.77 &                       1.58 \\
Random(2x)  &                       1.40 &                       0.70 &                       0.47 &                       0.36 &                       0.32 &                       2.99 &                       1.77 &                       1.44 &                       1.15 &                       1.02 \\
Random(4x)  &                       0.70 &                       0.36 &                       0.29 &                       0.24 &                       0.21 &                       1.77 &                       1.15 &                       0.89 &                       0.78 &                       0.69 \\
Random(10x) &                       0.32 &  \underline{\textbf{0.21}} &  \underline{\textbf{0.16}} &  \underline{\textbf{0.13}} &  \underline{\textbf{0.10}} &  \underline{\textbf{1.02}} &  \underline{\textbf{0.69}} &                       0.56 &                       0.51 &                       0.46 \\

\cmidrule(lr){2-6}
\cmidrule(lr){7-11}
\pruningwithrandomsearch{}  &                       0.96 &                       0.65 &                       0.50 &                       0.38 &                       0.32 &                       4.08 &                       2.30 &                       2.11 &                       1.72 &                       1.58 \\
\pruningwithBO{} &                       0.73 &              \textbf{0.27} &              \textbf{0.22} &              \textbf{0.19} &              \textbf{0.17} &                       5.10 &                       1.31 &                       0.68 &                       0.52 &                       0.45 \\
\smfo{}        &                       0.60 &              \textbf{0.32} &              \textbf{0.29} &              \textbf{0.27} &              \textbf{0.25} &                       3.71 &                       2.12 &                       1.98 &                       1.76 &                       1.74 \\
\wac{}         &                       5.14 &                       4.84 &                       4.82 &                       4.81 &                       4.81 &                       3.64 &                       3.30 &                       2.99 &                       2.90 &                       2.84 \\
\klweighting{}        &                       0.94 &                       0.68 &                       0.61 &                       0.28 &                       0.19 & - & - & - & - & - \\
\ablr{}        &                       4.09 &                       3.05 &                       2.98 &                       2.96 &                       2.95 &                       4.52 &                       2.46 &                       1.34 &                       0.93 &                       0.86 \\
\gcpplusprior{}   &                       2.98 &              \textbf{2.48} &              \textbf{1.69} &              \textbf{0.30} &              \textbf{0.26} &                       5.23 &                       4.08 &                       2.61 &                       2.24 &                       2.11 \\
\tstrei{}       &              \textbf{0.44} &              \textbf{0.38} &              \textbf{0.23} &              \textbf{0.23} &              \textbf{0.20} &              \textbf{1.35} &              \textbf{0.90} &                       0.64 &                       0.52 &                       0.49 \\
\taftstr{}  &                       0.36 &              \textbf{0.26} &              \textbf{0.21} &              \textbf{0.19} &              \textbf{0.17} &                       2.76 &                       1.15 &                       0.85 &                       0.60 &                       0.48 \\

\cmidrule(lr){2-6}
\cmidrule(lr){7-11}
\textbf{\tafrgpe{}}   &  \underline{\textbf{0.29}} &              \textbf{0.22} &              \textbf{0.19} &              \textbf{0.17} &              \textbf{0.16} &              \textbf{2.17} &              \textbf{1.78} &              \textbf{0.78} &              \textbf{0.48} &              \textbf{0.38} \\
\textbf{\rmogp{}}       &              \textbf{0.39} &              \textbf{0.23} &              \textbf{0.20} &              \textbf{0.18} &              \textbf{0.16} &                       2.33 &              \textbf{1.08} &                       0.74 &              \textbf{0.50} &              \textbf{0.30} \\
\textbf{\rgpemean{}}  &              \textbf{0.43} &              \textbf{0.31} &              \textbf{0.27} &              \textbf{0.25} &              \textbf{0.23} &                       1.78 &              \textbf{0.78} &  \underline{\textbf{0.45}} &  \underline{\textbf{0.38}} &  \underline{\textbf{0.26}} \\
\textbf{\rgpenei{}}         &                       3.29 &              \textbf{2.67} &              \textbf{2.22} &              \textbf{2.21} &              \textbf{2.20} &                       2.97 &              \textbf{0.91} &              \textbf{0.66} &              \textbf{0.44} &              \textbf{0.37} \\
\bottomrule
\end{tabular}
    \caption{Final results on the \glmnet{} and \svmfd{} benchmarks. The numbers reported are the average normalized regret~\citep{wistuba-ml18a}. We boldface the best value per benchmark and number of function evaluations and underline methods that are not significantly different according to a Wilcoxon signed-rank test with $\alpha=0.05$~\citep{demsar-06a}. Top group: standard HPO baselines; middle group: transfer HPO baselines; \textbf{bottom group: our methods}.\label{tab:results-glm-svm}}
\end{table}

\begin{table}[tb]
    \centering
    \small
\begin{tabular}{lllllllllll}
\toprule
{} & \multicolumn{1}{c}{10} &  \multicolumn{1}{c}{20} & \multicolumn{1}{c}{30} & \multicolumn{1}{c}{40} & \multicolumn{1}{c}{50} & \multicolumn{1}{c}{10} &  \multicolumn{1}{c}{20} & \multicolumn{1}{c}{30} & \multicolumn{1}{c}{40} & \multicolumn{1}{c}{50} \\
\midrule
{} & \multicolumn{5}{c}{\nn{}} & \multicolumn{5}{c}{\xgb{}} \\
\cmidrule(lr){2-6}
\cmidrule(lr){7-11}
GP     &                      16.40 &                      11.51 &                       9.69 &                       8.62 &                       7.84 &                       5.31 &                       2.36 &                       1.46 &                       1.22 &                       1.09 \\
GCP         &                      16.40 &                      11.61 &                       9.80 &                       8.56 &                       7.76 &                       5.31 &                       2.75 &                       2.10 &                       1.77 &                       1.50 \\
Random(1x)  &                      14.65 &                      11.50 &                       9.95 &                       8.75 &                       7.93 &                       5.68 &                       3.42 &                       2.47 &                       2.11 &                       1.80 \\
Random(2x)  &                      11.50 &                       8.75 &                       7.11 &                       6.13 &                       5.42 &                       3.42 &                       2.11 &                       1.62 &                       1.53 &                       1.36 \\
Random(4x)  &                       8.75 &                       6.13 &                       4.81 &                       4.24 &                       3.56 &                       2.11 &                       1.53 &                       1.31 &                       1.23 &                       1.17 \\
Random(10x) &                       5.42 &                       3.56 &              \textbf{2.62} &              \textbf{2.19} &              \textbf{1.78} &                       1.36 &                       1.17 &                       1.00 &                       0.92 &                       0.88 \\

\cmidrule(lr){2-6}
\cmidrule(lr){7-11}
\pruningwithrandomsearch{}  &                      12.73 &                       9.78 &                       8.38 &                       7.23 &                       6.51 &                       3.92 &                       2.60 &                       1.62 &                       1.53 &                       1.43 \\
\pruningwithBO{} &                      11.79 &                       8.85 &                       7.08 &                       6.06 &                       5.14 &                       2.75 &                       1.18 &                       0.89 &                       0.78 &                       0.73 \\
\smfo{}        &              \textbf{3.59} &              \textbf{2.77} &              \textbf{2.17} &              \textbf{1.86} &              \textbf{1.75} &                       1.25 &                       1.09 &                       0.94 &                       0.91 &                       0.87 \\
\wac{}         &                       7.44 &                       6.68 &                       5.98 &                       5.79 &                       5.51 &                       1.88 &                       1.85 &                       1.84 &                       1.75 &                       1.75 \\
\klweighting{}        &                       6.43 &                       5.44 &                       4.67 &                       4.17 &                       3.78 & - & - & - & - & - \\
\ablr{}        &                       7.88 &                       6.60 &                       6.02 &                       5.45 &                       5.03 &                       2.29 &                       1.88 &                       1.75 &                       1.61 &                       1.58 \\
\gcpplusprior{}   &                       5.93 &                       4.30 &                       3.28 &                       2.74 &                       2.38 &                       1.37 &              \textbf{1.18} &              \textbf{1.08} &              \textbf{0.94} &              \textbf{0.84} \\
\tstrei{}       &                       5.37 &                       4.46 &                       3.88 &                       3.31 &                       3.00 &              \textbf{1.14} &              \textbf{0.88} &                       0.84 &              \textbf{0.79} &                       0.78 \\
\taftstr{}  &                       4.88 &                       4.02 &                       3.58 &                       3.24 &                       2.95 &              \textbf{0.98} &              \textbf{0.82} &              \textbf{0.73} &              \textbf{0.69} &  \underline{\textbf{0.64}} \\

\cmidrule(lr){2-6}
\cmidrule(lr){7-11}
\textbf{\tafrgpe{}}   &              \textbf{3.60} &  \underline{\textbf{2.54}} &  \underline{\textbf{2.09}} &  \underline{\textbf{1.70}} &  \underline{\textbf{1.55}} &  \underline{\textbf{0.97}} &              \textbf{0.80} &              \textbf{0.70} &              \textbf{0.67} &              \textbf{0.67} \\
\textbf{\rmogp{}}       &  \underline{\textbf{3.52}} &              \textbf{2.63} &                       2.34 &                       2.07 &                       1.93 &                       1.55 &                       1.16 &                       0.89 &                       0.83 &                       0.80 \\
\textbf{\rgpemean{}}  &                       5.20 &                       3.92 &                       3.29 &                       2.91 &                       2.55 &              \textbf{1.12} &  \underline{\textbf{0.79}} &  \underline{\textbf{0.68}} &  \underline{\textbf{0.67}} &              \textbf{0.65} \\
\textbf{\rgpenei{}}         &                       5.09 &                       3.90 &                       3.38 &                       3.00 &                       2.71 &                       1.19 &              \textbf{0.98} &              \textbf{0.87} &              \textbf{0.73} &              \textbf{0.64} \\

\bottomrule
\end{tabular}
    \caption{Final results on the \nn{} and \xgb{} benchmarks. The numbers reported are the average normalized regret~\citep{wistuba-ml18a}. We boldface the best value per benchmark and number of function evaluations and underline methods that are not significantly different according to a Wilcoxon signed-rank test with $\alpha=0.05$~\citep{demsar-06a}. Top group: standard HPO baselines; middle group: transfer HPO baselines; \textbf{bottom group: our methods}.\label{tab:results-nn-xgb}}
\end{table}

\begin{table}[tb]
    \centering
    \small
\begin{tabular}{lllllllllll}
\toprule
{} & \multicolumn{1}{c}{10} &  \multicolumn{1}{c}{20} & \multicolumn{1}{c}{30} & \multicolumn{1}{c}{40} & \multicolumn{1}{c}{50} & \multicolumn{1}{c}{10} &  \multicolumn{1}{c}{20} & \multicolumn{1}{c}{30} & \multicolumn{1}{c}{40} & \multicolumn{1}{c}{50} \\
\midrule
{} & \multicolumn{5}{c}{\adaboost{}} & \multicolumn{5}{c}{\svmsd{}} \\
\cmidrule(lr){2-6}
\cmidrule(lr){7-11}
GP     &                       5.42 &              \textbf{2.26} &              \textbf{1.26} &              \textbf{0.82} &              \textbf{0.66} &                       9.66 &                       3.64 &                       2.06 &                       1.43 &                       1.13 \\
GCP         &                       5.42 &  \underline{\textbf{2.26}} &  \underline{\textbf{1.18}} &              \textbf{0.92} &                       0.72 &                       9.66 &                       3.59 &                       1.97 &                       1.18 &                       0.81 \\
Random(1x)  &                       6.27 &                       3.62 &                       2.53 &                       1.72 &                       1.32 &                      11.52 &                       6.44 &                       5.07 &                       4.07 &                       3.24 \\
\cmidrule(lr){2-6}
\cmidrule(lr){7-11}
\pruningwithrandomsearch{}  &                       5.43 &                       3.61 &                       2.37 &                       1.79 &                       1.45 &                      10.64 &                       6.09 &                       4.26 &                       3.37 &                       2.89 \\
\pruningwithBO{} &                       5.38 &              \textbf{2.31} &              \textbf{1.27} &  \underline{\textbf{0.81}} &              \textbf{0.72} &                       9.86 &                       3.53 &                       2.02 &                       1.48 &                       1.18 \\
\smfo{}        &              \textbf{4.03} &                       2.76 &                       1.96 &                       1.52 &                       1.11 &                       4.30 &                       2.63 &                       1.97 &                       1.37 &                       1.21 \\
\wac{}         &                       5.97 &                       4.11 &                       3.01 &                       2.18 &                       1.42 &                       8.49 &                       6.02 &                       3.67 &                       2.63 &                       1.90 \\
\klweighting{}        &                       5.49 &                       3.48 &                       2.73 &                       2.23 &                       1.81 \\
\ablr{}        &                       4.65 &                       2.39 &                       1.51 &                       0.88 &                       0.52 &                       7.65 &                       4.84 &                       3.39 &                       2.47 &                       1.86 \\
\gcpplusprior{}   &              \textbf{5.37} &              \textbf{3.38} &              \textbf{1.92} &              \textbf{1.25} &                       0.89 &              \textbf{4.23} &              \textbf{2.59} &              \textbf{2.02} &              \textbf{1.85} &              \textbf{1.61} \\
\tstrei{}       &                       4.75 &              \textbf{2.59} &              \textbf{1.65} &                       0.94 &  \underline{\textbf{0.52}} &                       3.69 &                       2.07 &                       1.36 &                       0.89 &              \textbf{0.64} \\
\taftstr{}  &              \textbf{3.96} &              \textbf{2.35} &              \textbf{1.54} &                       1.14 &              \textbf{0.73} &                       3.89 &                       2.24 &                       1.39 &                       0.78 &              \textbf{0.46} \\

\cmidrule(lr){2-6}
\cmidrule(lr){7-11}
\textbf{\tafrgpe{}}   &  \underline{\textbf{3.91}} &              \textbf{2.29} &              \textbf{1.39} &              \textbf{1.00} &              \textbf{0.63} &  \underline{\textbf{2.95}} &              \textbf{1.54} &                       0.91 &  \underline{\textbf{0.61}} &              \textbf{0.45} \\
\textbf{\rmogp{}}       &              \textbf{4.01} &              \textbf{2.26} &              \textbf{1.47} &              \textbf{0.95} &              \textbf{0.76} &                       3.35 &                       2.04 &                       1.35 &              \textbf{0.92} &              \textbf{0.58} \\
\textbf{\rgpemean{}}  &                       4.71 &              \textbf{2.48} &              \textbf{1.63} &              \textbf{0.94} &              \textbf{0.55} &              \textbf{3.22} &              \textbf{1.70} &              \textbf{0.99} &              \textbf{0.71} &              \textbf{0.47} \\
\textbf{\rgpenei{}}         &                       4.55 &              \textbf{2.58} &                       1.76 &                       1.09 &              \textbf{0.62} &              \textbf{3.75} &  \underline{\textbf{1.41}} &  \underline{\textbf{0.75}} &              \textbf{0.61} &  \underline{\textbf{0.39}} \\

\bottomrule
\end{tabular}
    \caption{Final results on the \adaboost{} and \svmsd{} benchmarks. The numbers reported are the average normalized regret~\citep{wistuba-ml18a}. We boldface the best value per benchmark and number of function evaluations and underline methods that are not significantly different according to a Wilcoxon signed-rank test with $\alpha=0.05$~\citep{demsar-06a}. Top group: standard HPO baselines; middle group: transfer HPO baselines; \textbf{bottom group: our methods}.\label{tab:results-ada-svm}}
\end{table}

Our proposed methods consistently outperform the standard HPO baselines vanilla BO and random search. Comparing our proposed method to the \tstrei{} baseline, we find that \tstrei{} can perform well on some tasks, but fails to do so on the \svmfd{} and \nn{} benchmarks. We find that our methods consistently outperform the additional baselines \wac{}, \ablr{} and \klweighting{}.  We find these baselines to have unreliable performance and suggest further research into their robustness. \gcpplusprior{} showed good performance on some problems, but our proposed methods yield further improvements in mean performance on all six benchmarks, and significant improvements on three of them.

The \smfo{} baseline is a stronger baseline than the regular GP and is particularly strong on the \nn{} benchmark, where it outperforms all methods except for \tafrgpe{}. The small difference in performance between the GP and random search shows that modeling provides limited benefit on this benchmark. However, it is outperformed on the other five benchmarks by most transfer learning methods. 
We would like to note that the \smfo{} baseline outperforming the proposed methods on a single benchmark is not a drawback~\citep{sculley-iclrws18a} as one would not know that the baseline performs well on this benchmark in advance, and using the baseline on all benchmarks would lead to clearly inferior results.

The search space pruning baselines (\emph{\pruningwithrandomsearch{}} and \emph{\pruningwithBO}) show that shrinking the search space indeed improves performance. However, shrinking the search space does not simplify the search space and therefore does not improve the GP much on the \nn{} benchmark. Nonetheless, there is still a performance improvement on this benchmark, and the reduced search space appears to be a good idea. We further find that search space pruning is consistently outperformed on the table-lookup benchmarks (\adaboost{}, \svmsd{} and \nn{}) while it performed competitively on the surrogate benchmarks. This hints at the two types of benchmarks (tabular vs. surrogates) having a different underlying structure, which is an interesting topic for further studies.

\begin{figure}
    \centering
    \includegraphics[width=0.99\textwidth]{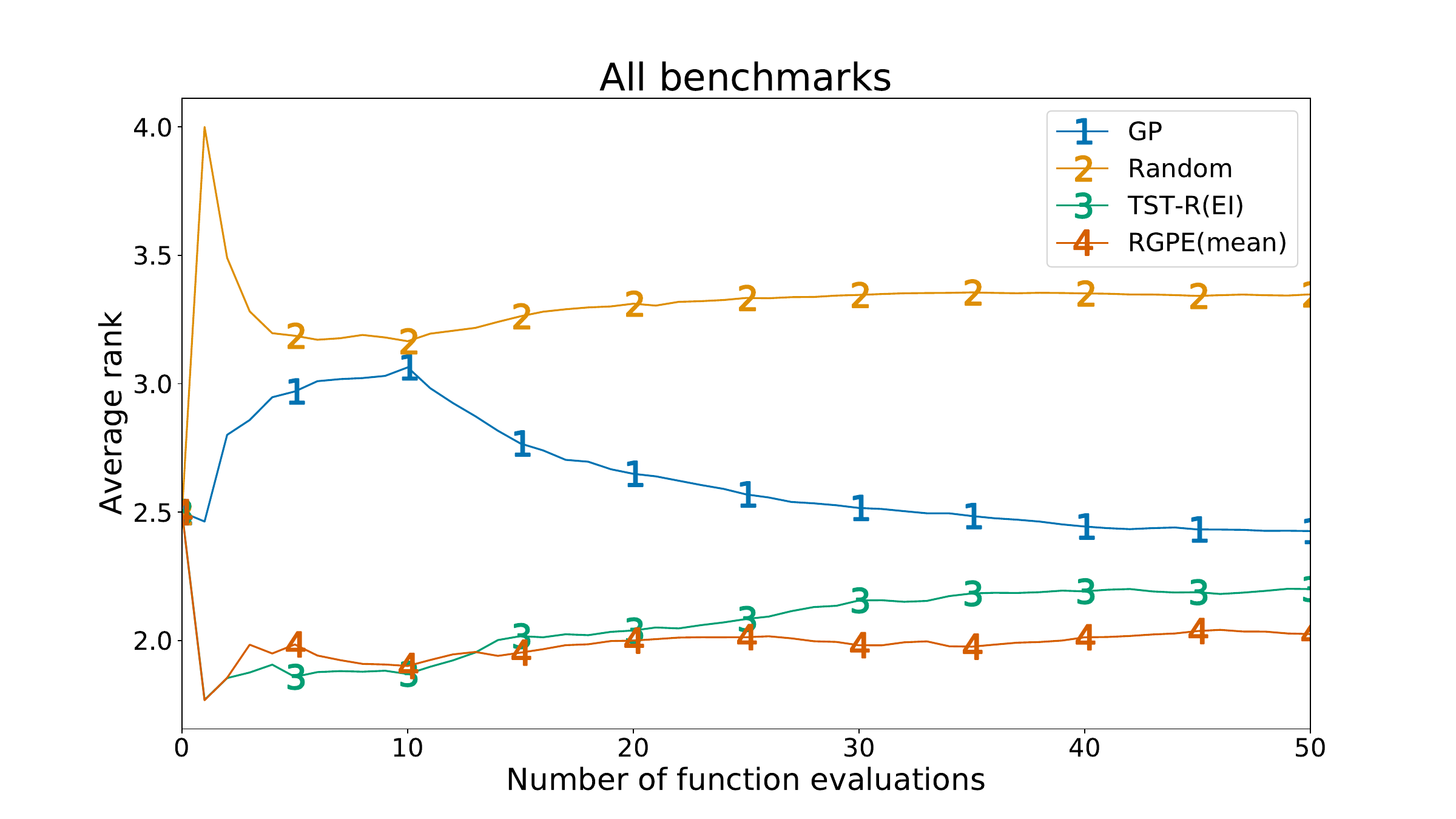}
    \includegraphics[width=0.99\textwidth]{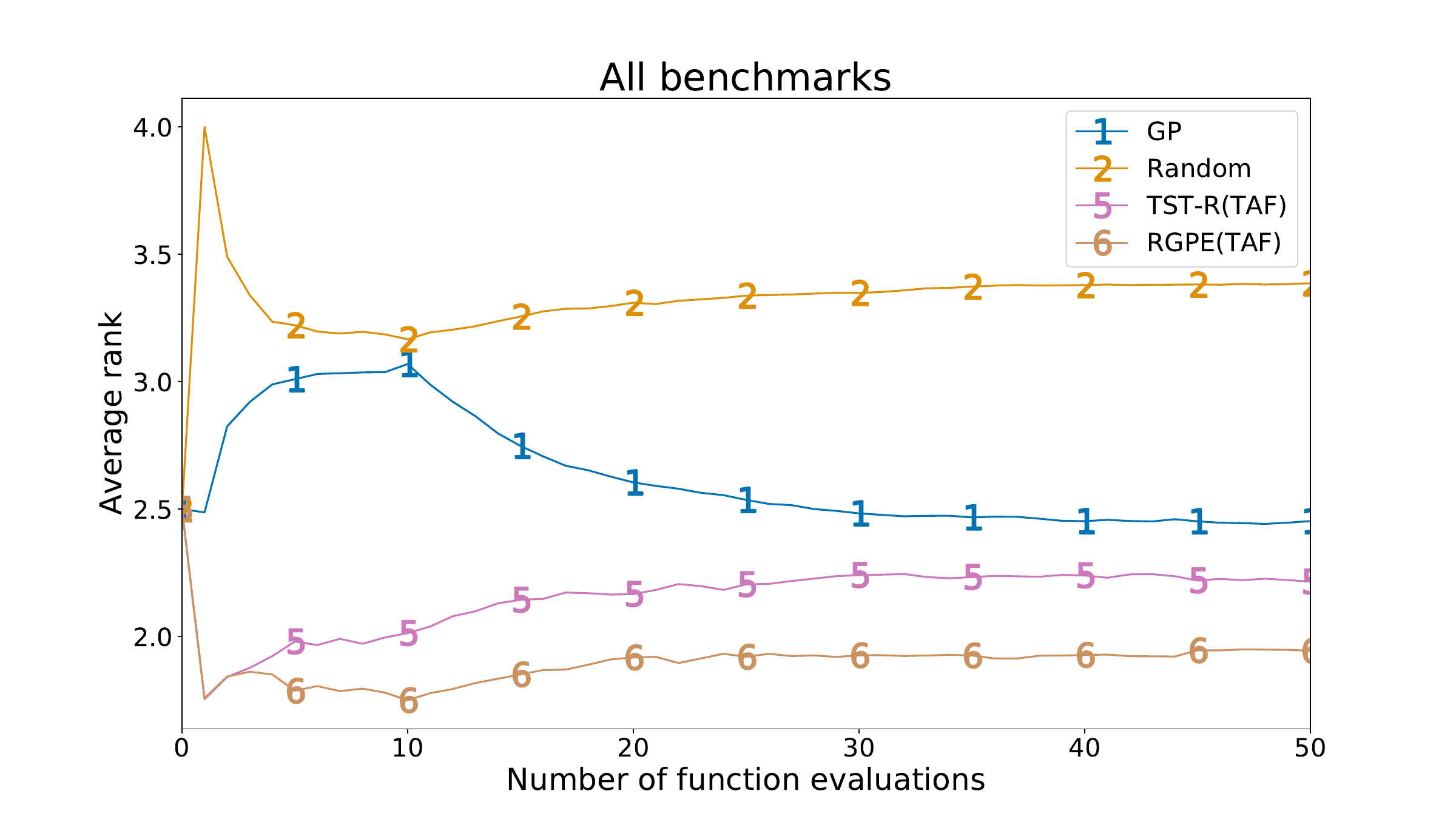}
    \caption{Ranking plots across all benchmarks (lower is better) comparing our methods (4,6) with small sets of baselines (1-3,5), as indicated in each panel legend. Our methods consistently outperformed baselines across all benchmarks.}
    \label{fig:ranking}
\end{figure}

Figure~\ref{fig:ranking} provides a graphical representation of the results comparing our best method based on a single ensemble (\rgpemean{}) and our best method based on an improved acquisition function (\tafrgpe{}) against their respective baselines (\tstrei{} and \taftstr{}). We present the baselines GP and random search together with \tstrei{} and \rgpemean{} (top), and \taftstr{} and \tafrgpe{} (bottom) . We can observe all transfer HPO methods benefit from the learned initial design immediately, while the GP using the Latin hypercube initial design and random search cannot keep up. The ranks of the GP and random search appear to converge towards each other until iteration 10, where the GP's initial design is over and BO actually starts, and from that moment on the GP clearly outperforms random search. The transfer HPO methods maintain their lead over the single-task methods over the course of the 50 function evaluations. We can observe that our proposed methods outperform the respective competitors without requiring any hyperparameter optimization (in contrast to the methods based on \tstr{}).

\subsection{Evolution of Model Weights}

\begin{figure}[tb]
    \centering
    \includegraphics[width=0.99\textwidth]{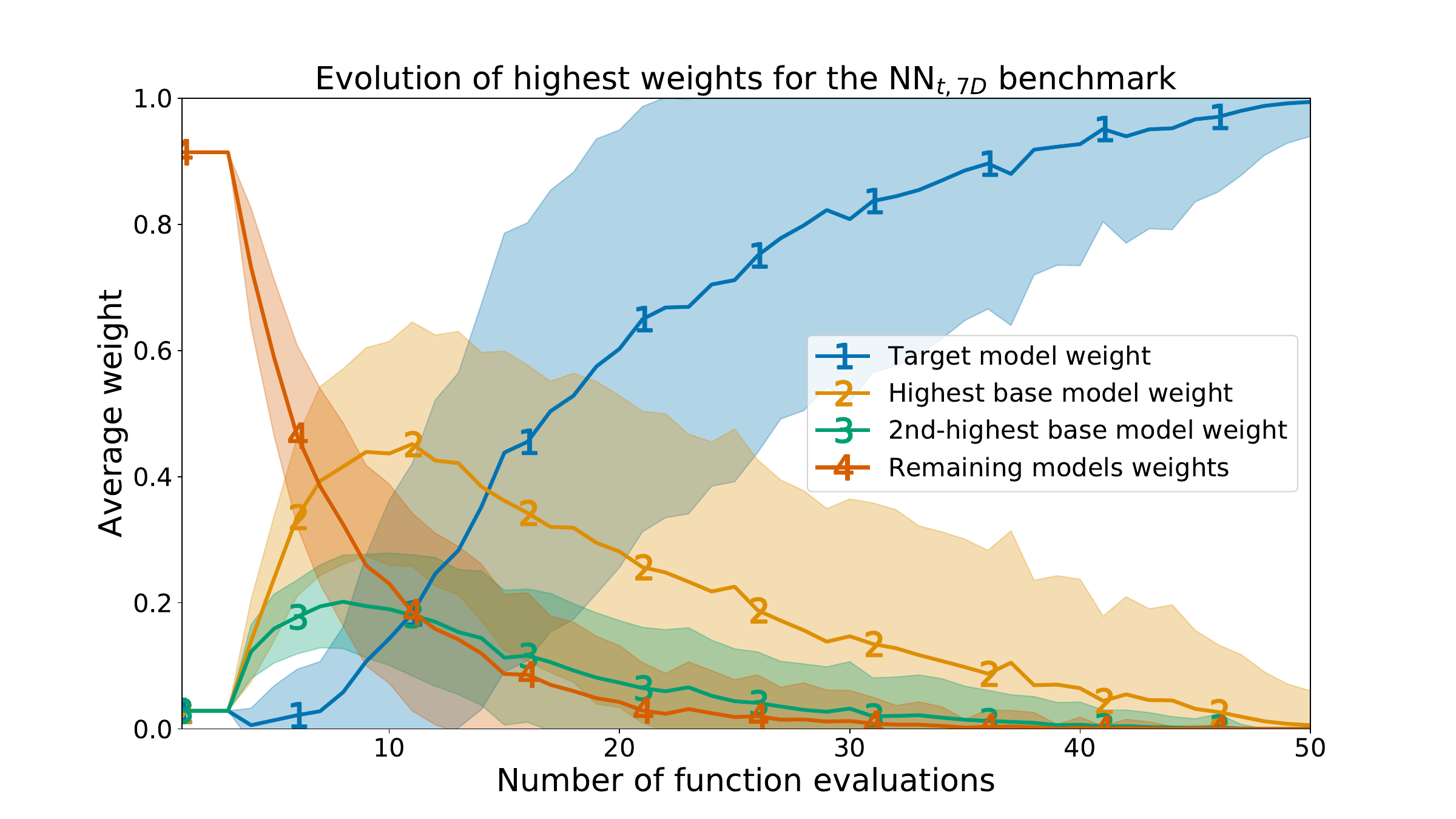}
    \includegraphics[width=0.99\textwidth]{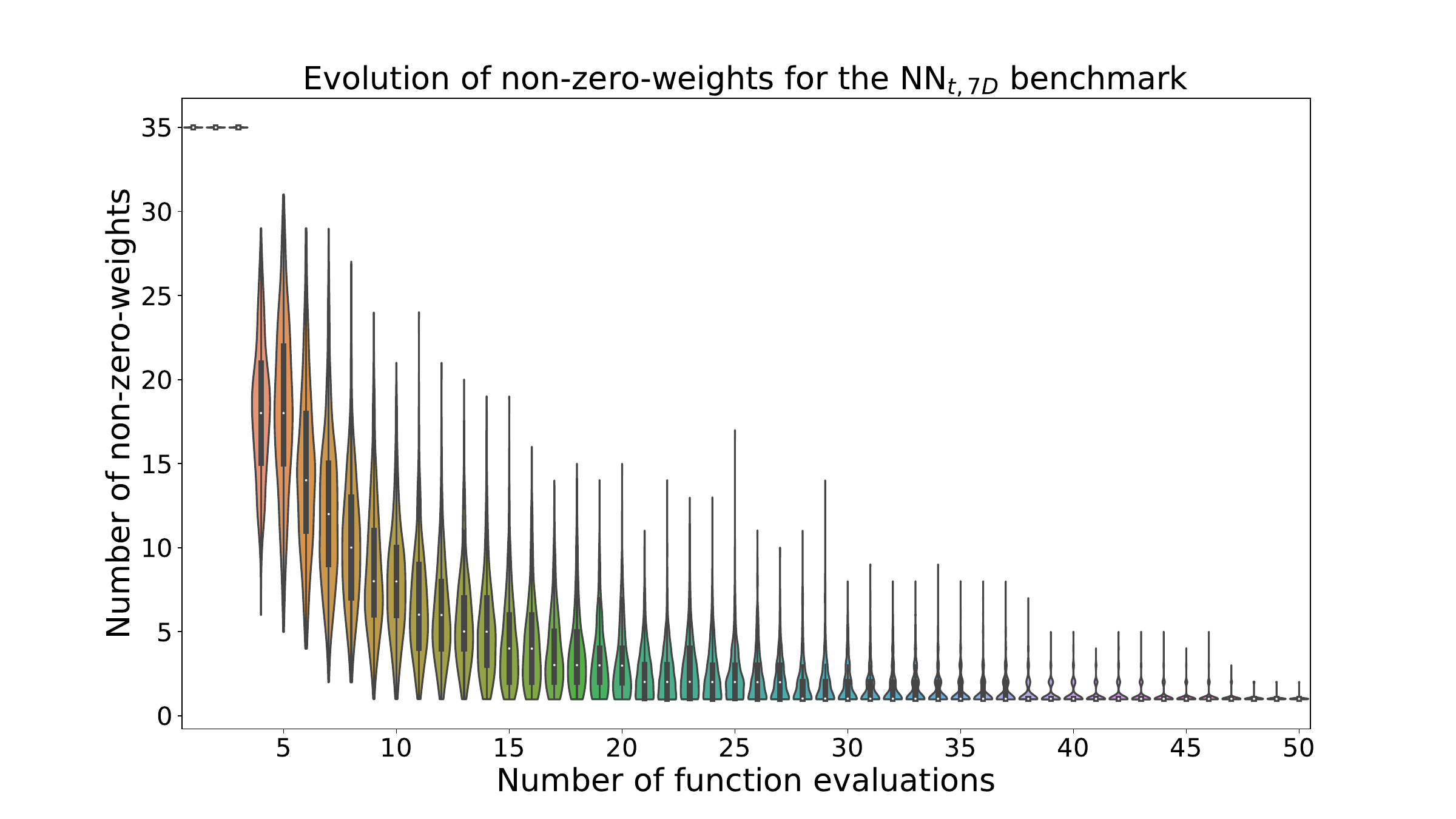}
    \caption{(Top) Evolution of the individual model weights over time for the \rgpenei{} model. (Bottom) Evolution of the number of non-zero weights over time for the \rgpenei{} model.}
    \label{fig:weights}
\end{figure}

Figure~\ref{fig:weights} shows how the ensemble weights evolved for the \rgpenei{} method on the \nn{} benchmark. The top panel depicts how the weight of the target model increases over the course of the 50 function evaluations and its relation to the weights of the base models. This plot shows the average of the weight over all 15 repetitions and 35 tasks. For the highest and 2nd highest weight of a base model, we identify it per time step before taking the average. We can see that the weight of the target model is initially close to zero before rapidly approaching a weight close to 1 by the final iteration. In early iterations the weight was spread across all of the base models, but with more data, weight quickly concentrated on a small number of base models, before finally concentrating on the target model. By iteration 11 the two most highly-weighted base models already had more weight than all of the other base models combined. On the bottom we can see the evolution of non-zero weights for the \nn{} benchmark. For the first three iterations there are none, as the proposed weighting scheme requires three function evaluations before it can be used. As soon as the weighting kicks in, half of the weights are set to zero (in the median). The number of non-zero weights decreases monotonically and the median approaches one (all weight on the target model) at 28 function evaluations.

\subsection{Ablation Study}
\label{sec:ablation}

In this second part of the experimental validation we examine the effect of the proposed building blocks of our method on the performance. We conduct an ablation study and sensitivity analaysis for the \tstrei{} and \taftstr{} baselines in Appendix~\ref{ssec:ablation_tstr_taf}.

\begin{table}[tb]
    \centering
    \small

\begin{tabular}{
l
p{0.11\textwidth}@{\hskip 0mm}
p{0.11\textwidth}@{\hskip 0mm}
p{0.11\textwidth}@{\hskip 0mm}
p{0.11\textwidth}@{\hskip 0mm}
p{0.11\textwidth}@{\hskip 0mm}
p{0.11\textwidth}
}
\toprule
{} & \rotatebox[origin=b]{50}{\glmnet{}} & \rotatebox[origin=b]{50}{\svmfd{}} & \rotatebox[origin=b]{50}{\nn{}} & \rotatebox[origin=b]{50}{\xgb{}} & \rotatebox[origin=b]{50}{\adaboost{}} & \rotatebox[origin=b]{50}{\svmsd{}} \\
\midrule
\rgpemean{} &              \textbf{0.23} &  \underline{\textbf{0.26}} &  \underline{\textbf{2.55}} &              \textbf{0.65} &  \underline{\textbf{0.55}} &              \textbf{0.47} \\
\rgpeei{}         &              \textbf{0.21} &              \textbf{0.31} &              \textbf{2.69} &              \textbf{0.67} &              \textbf{0.57} &              \textbf{0.46} \\
\rgpenei{}        &              \textbf{2.20} &              \textbf{0.37} &              \textbf{2.71} &  \underline{\textbf{0.64}} &              \textbf{0.62} &  \underline{\textbf{0.39}} \\
\rgpenowd{}  &  \underline{\textbf{0.16}} &              \textbf{0.48} &                       3.27 &              \textbf{0.71} &              \textbf{0.68} &              \textbf{0.45} \\
\midrule
\tafrgpe{}      &                       0.16 &              \textbf{0.38} &  \underline{\textbf{1.55}} &              \textbf{0.67} &              \textbf{0.63} &  \underline{\textbf{0.45}} \\
\tafrgpenowd{} &              \textbf{0.15} &              \textbf{0.51} &              \textbf{1.61} &  \underline{\textbf{0.62}} &                       0.71 &              \textbf{0.45} \\
\rmogp{}          &              \textbf{0.16} &  \underline{\textbf{0.30}} &                       1.93 &                       0.80 &              \textbf{0.76} &              \textbf{0.58} \\
\rmogpnowd{}    &  \underline{\textbf{0.11}} &              \textbf{0.56} &                       1.96 &                       0.75 &  \underline{\textbf{0.61}} &                       0.58 \\
\midrule
\rmogponehundred{}    &              \textbf{0.15} &                       0.55 &                       2.00 &              \textbf{0.72} &              \textbf{0.64} &  \underline{\textbf{0.49}} \\
\rmogponethousand{}             &              \textbf{0.16} &  \underline{\textbf{0.30}} &              \textbf{1.93} &              \textbf{0.80} &              \textbf{0.76} &              \textbf{0.58} \\
\rmogptenthousand{}  &  \underline{\textbf{0.14}} &              \textbf{0.37} &              \textbf{1.90} &  \underline{\textbf{0.68}} &                       0.58 &              \textbf{0.54} \\
\rmogponehundredthousand{} &              \textbf{0.17} &              \textbf{0.31} &  \underline{\textbf{1.83}} &              \textbf{0.73} &  \underline{\textbf{0.40}} &              \textbf{0.50} \\
\bottomrule
\end{tabular}

    \caption{Ablation study on the proposed methods. The numbers reported are the average normalized regret~\citep{wistuba-ml18a}. We boldface the best value per benchmark and number of function evaluations and underline methods that are not significantly different according to a Wilcoxon signed-rank test with $\alpha=0.05$~\citep{demsar-06a}; separately for \rgpenei{}, \tafrgpe{} and \rmogp{} with different values of $S$.}
    \label{tab:ablation}

\end{table}

\begin{table}[tb]
    \centering
    \begin{tabular}{lcccc}
    \toprule
        Method name & 4.1 & 4.3 & Acquisition function \\
        \midrule
        \tstrei{} & {} & {} & EI   \\
        \tstrwd{}, see Appendix~\ref{ssec:ablation_tstr_taf} & {} & $\checkmark$ & EI  \\
        \rgpenei{} & $\checkmark$ & $\checkmark$ & NoisyEI \\
        \rgpenowd{} & $\checkmark$ & {} & NoisyEI  \\
        \rgpemean{} & $\checkmark$ & $\checkmark$ & EI  \\
        \rgpeei{}  & $\checkmark$ & $\checkmark$ & EI  \\
        \taftstr{} & {} & {} & TAF  \\
        \taftstrwd{}, see Appendix~\ref{ssec:ablation_tstr_taf} & {} & $\checkmark$ & TAF  \\
        \tafrgpe{} & $\checkmark$ & $\checkmark$ & TAF  \\
        \tafrgpenowd{} & $\checkmark$ & {} & TAF  \\
        \rmogp{} & $\checkmark$ & $\checkmark$ & Weighted sum of EIs \\
        \rmogp{}(None) & $\checkmark$ & {} & Weighted sum of EIs \\
        \bottomrule
    \end{tabular}
    
    \caption{Overview of how our proposed methods and their ablations compare against the baselines \tstrei{} (line 1) and \taftstr{} (line 6).}
    \label{tab:abbrevations}
\end{table}

In the methodology section we have introduced the \rgpe{} model and how it can be used for transfer HPO. Because the weighting mechanism contains a regularization mechanism that can be turned off, we conduct an ablation study about it. Moreover, it contains a hyperparameter (the number of bootstrap samples) of which we study the sensitivity. We present all results in Table~\ref{tab:ablation}. Furthermore, we provide an overview of how the method names and the exact contributions discussed in Section~\ref{sec:methodology} correspond in Table~\ref{tab:abbrevations}.

First, we check whether using the weight dilution prevention is a useful addition, and we can compare \rgpenei{} and \rgpenowd{}, \tafrgpe{} and \tafrgpenowd{}, and \rmogp{} and \rmogp{}(None). We can see that there is no large difference between using and not using the weight dilution mechanism, and a slight trend toward the methods with weight dilution prevention achieving lower regret values. However, there are two outliers. \rgpenei{} with the weight dilution prevention has a drastically worse performance than \rgpenowd{} on \glmnet{}, but this is not significant and caused by a single dataset. On the other hand, \rgpenowd{} is significantly worse than \rgpenei{} on \nn{}. Therefore, we can conclude that using the weight dilution prevention, which is also central to our proof that the proposed methods are in the worst case only a multiplicative factor slower than vanilla Bayesian optimization and improves optimization speed by more aggressively dropping base models, is a valuable contribution for methods combining multiple models into an ensemble.

Second, we analyze whether the methods that make use of the predictive variance in the base models have an edge over those that do not. This time we compare \rgpenei{} and \rgpemean{}, and \rmogp{} and \tafrgpe{}. Surprisingly, even though we have shown in a principled manner why the variance of the base models should matter, ignoring the variance of the base models as done by \rgpemean{} and \tafrgpe{} gives better results. While the difference in results for \rgpemean{} and \rgpenei{} are not statistically significant, \rgpemean{} has lower regret on four out of six benchmarks and the lowest regret among the four compared methods on three out of six benchmarks. The situation for \rmogp{} and \tafrgpe{} is slightly more nuanced, but similar overall. Our best explanations for this surprising result are 1) that predictive variance in base models can be misleading and increase the supposed variance of the combined model in regions where the target model is already certain, thereby misleading Bayesian optimization; and 2) that as there is no variance around already evaluated points in base models, these points (including the optima on the base models) appear less promising after combining all base models, repelling the transfer HPO methods from such locations.

Third, we discuss whether the special treatment of noise in the \rgpenei{} model is actually necessary or whether plain expected improvement would work, too (\rgpeei{}, line 2 of Table~\ref{tab:ablation}). It turns out that there is no statistically significant degradation of the performance. We can currently not explain why there is no significant difference in the way we treat the uncertainty of the base model and suggest checking this finding when using the \rgpe{} model with a different GP library. 

Fourth, we have a look at using a different number of bootstrap samples than the default $1000$, and also test $100$, $10000$ and $100000$ bootstrap samples. Using as many as $100000$ appears to perform consistently better or equivalent to using only $1000$ samples. Contrarily, using only $100$ samples results in significantly worse performance in two out of six cases. Therefore, we do not see a reason to tune the number of bootstrap samples but rather use the maximum number of samples that can be afforded. As a concrete example, it takes less than roughly 1.6 seconds to compute the weights with $1000$ bootstrap samples after 50 observations on a $2d$ problem, while it takes 140 seconds for $100000$ bootstrap samples. This increase in computational requirements would not have been feasible for the large experimental study we conducted, but can be used in practical settings.

Overall, we find our proposed methods to be very robust. We see some advantage to ignoring the predictive variance of the base models and using the weight dilution prevention. Based on the ablation study and the overall results in the previous section, our clear recommendation is to use \tafrgpe{} for transfer HPO.

\section{Related Work}\label{app:related}

This section extends the primer on related work in Section 2.3 by discussing techniques of types 1-5 and 7-9 which are not required as background for our newly developed methods.

\paragraph{1. A single model that is trained on all tasks simultaneously.}
Several past methods have used manually defined meta-features to measure task similarity \citep{brazdil-ecml94a} and then adapted GP-based Bayesian optimization \citep{bardenet-icml13a,yogatama-aistats14a,schilling-pkdd16a}. Besides the drawback of requiring additional hand-designed features (which violates desideratum \#3), these methods share the issue that meta-features are non-adaptive throughout the optimization process and do not update task similarity based on the new observations~\citep{leite-mldmpr12,wistuba-ml18a}.

Another set of methods attempts to learn the task similarity without the use of meta-features in order to fit a joint model. \citet{swersky-nips13a} and \citet{poloczek-wsc16a} propose different kernels to jointly model all past runs and the current task. However, multitask GPs suffer from the same poor scaling as putting all observations into a single GP (violating desideratum \#1) and cannot be used for the benchmarks we use in the main paper. Furthermore, \citet{swersky-nips13a} sample a $t \times t$ lower triangular matrix describing task correlations, which prohibits a large number of past runs as observed by \citet{klein-aistats17}. We demonstrate this scaling issue in Appendix~\ref{sec:methodology:scaling} where one can see that the fitting time for the multitask GP with only five tasks is already higher than that of \rgpe{} for 50 tasks, and that fitting with 15 tasks already requires more than 1000 seconds.

Such scaling issues can be alleviated by using neural network models. One option is to learn a task embedding for every task inside a Bayesian neural network~\citep{springenberg-nips16a} and another option is to fine-tune a pre-trained deep kernel surrogate on the target task~\citep{wistuba-iclr21a}.
Two recent works~\citep{perrone-neurips18a,law-neurips19a} circumvent both the scaling issue as well as the issue of explicitly learning task correlations by using a neural network as a feature extractor for tasks and hyperparameters and Bayesian linear regression as a scalable probabilistic output layer of the neural network. We compared to \emph{adaptive basis function linear regression} ~\citep[\ablr{}]{perrone-neurips18a} in Section~\ref{sec:experiments}.

When the hyperparameter search space is discrete it is also possible to use methods based on a matrix factorization model~\citep{fusi-neurips18a,yang-kdd2019a,yang-kdd2020a}. However, these methods require that there are overlapping observations between the datasets and were also only tested in settings where the metadata matrix (i.e. results of pipelines on all datasets) has between 10\% of the entries~\citep{fusi-neurips18a} or is even fully populated. Such approaches are related to algorithm selection, where the goal is to select a single algorithm from a discrete set without observing any feedback on the target task~\citep{kerschke-evocomp2019a}, see also the method we discuss in Appendix~\ref{app:active_testing}.

\paragraph{2. A learned kernel that is trained on the base tasks and applied to the target task.}
As an alternative solution, \citet{wang-neurips2018a} suggest to estimate the Gaussian process prior from offline data via empirical Bayes. While the learned kernel leads to performance improvements compared to a standard RBF prior, the size of the meta dataset required for the method to perform en-par with the RBF prior with optimized hyperparameters is between 
2916 
and 20000, 
which is above what we will have available during the experimental evaluation.

\paragraph{3. Learning an adaptation of each base task to the current task}
A number of papers propose to learn an adaptation of each past BO run to the current BO run. \citet{schilling-pkdd16a} learn a joint model for each past run and the target task and combine those models using meta-features as similarity descriptors in a Product of Gaussian Process Experts model. \citet{shilton-aistats17a} model the difference between the past run and the target task with a GP, which is then used to adjust the past run observations for inclusion in the current model. Such approaches are not scalable as they require re-fitting a model for each meta-task in each iteration of the optimization process (violating desideratum \#2). \citet{ramachandran-pricai18a} then suggest a bandit method to select a base task to update within this transfer setting. Due to the necessary exploration this method is only applicable in the setting where the number of base tasks is substantially lower than the number of observations allowed on the target task (violating desideratum \#1). \citet{golovin-kdd17a} take a similar approach for an ordered set of past runs. Rather than fitting a GP to each run separately, a GP is fit to the residuals of each run relative to the predictions of the previous model in the stack. However, this method assumes an ordering of the runs, which would not be the case for transferring information from a collection of unrelated problems.

\paragraph{4. An initial design learned from previous optimization runs}
Orthogonal to optimization is learning the initial design based on past observations, which can be done both with meta-features~\citep{reif-ml12a,gomes-2012a,feurer-aaai15a} and without~\citep{wistuba-icdm2015a,pfisterer-gecco21a}. Using an adapted initial design is complementary to our strategy and would benefit our proposed method as well.
We compared against an initial design-only baseline, also called \emph{sequential model-free optimization}~\citep[\smfo{}]{xu-aaai10a,xu-rcra11a,wistuba-icdm2015a,pfisterer-gecco21a}, in Section~\ref{sec:experiments}.

\paragraph{5. Reducing the design space based on previous optimization runs}
Yet another orthogonal strategy is to use past optimization runs to shrink the considered search space to only contain promising areas from past runs~\citep{wistuba-pkdd15a,probst-jmlr2019a,perrone-neurips19a}.  However, most of these strategies introduce at least one new hyperparameter (violating desideratum \#4). The only exception is given by \citet{perrone-neurips19a}, which also introduces a hyperparameter-free method to reduce the search space to a box which only contains all prior observed optima. We compared to this technique in Section~\ref{sec:experiments} (\pruningwithrandomsearch{},\pruningwithBO).

\paragraph{7. Linear combination of models} In the related field of algorithm configuration~\citep{Hutter09}, \citet{lindauer-aaai18} use stochastic gradient descent to learn a weighted combination of random forests as surrogates for a runtime optimization task.
The mean predictions are combined linearly as in \tstr{}~\citep{wistuba-ecml16a}, and the variances are combined according to $\bar{\sigma}^2(\mathbf{x}_*) = \sum_{i=1}^t w_i\sigma^2_i(\mathbf{x}_*)$. 
While the usage of a simple stacking approach is appealing, it introduces a new model selection problem to find a stacking regressor (violating desideratum \#3). 
Without proper tuning, obtaining a sparse solution (sparse in terms of taking only relevant base tasks into account while ignoring unrelated ones) is not possible.
Furthermore, minimizing the least squares error is not necessarily a good criterion when our goal is to obtain a good model for Bayesian optimization. We compared against this stacking approach abbreviated \emph{\wac{}} in Section~\ref{sec:experiments}.

Another line of work extends the Predictive Entropy Search (PES) acquisition function to a multi-task Bayesian optimization algorithm in which the distribution over the minimum is a weighted linear combination of the distribution of the minimum on different tasks~\citep{ramachandran-pkdd19a}. The linear weighting is computed as $\exp{( -\frac{KL(p^{base}_{opt}(\mathbf{x})||p^{target}_{opt}(\mathbf{x}))}{\eta} )}$, where $\eta$ is a hyperparameter and $KL$ the Kullback-Leibler divergence estimated with a nearest-neighbor approach~\cite{perez-cruz-ieee08a}. Later, they also proposed a modification that takes uncertainty on the source task into account at the cost of yet an additional hyperparameter \citep{ramachandran-aajcai19a}.
While here, we restrict ourselves to EI-based acquisition functions, we note that our proposed weighting scheme could also be used as a hyperparameter-free alternative in PES.

\paragraph{8. Hybrid methods} While the ideas above can be categorized rather easily, it is also possible to define models and techniques that are combinations of the methods described above. One such method is to fit a joint neural network to all base tasks without taking the tasks into account for the network architecture to learn a global prior. To take the target task into account, this method uses a GP that only learns the differences of the current task to the global prior. To ensure all tasks live on the same scale, the data is scaled using copula transformations. As this changes the predictive distribution of the GP, they also derive a custom, closed-form solution for EI tailored to this model. We compare against this method \emph{Copula Gaussian processes with prior}, or short \gcpplusprior{}~\citep{salinas-icml2020a}, in the experimental section. Another combination involves meta-features and a learning a data-dependent default~\citep{gijsbers-gecco21a}.

\paragraph{9. Prior-based methods} A transfer that we did not discussed yet is inferring a prior from previous task and applying to existing methods. An obvious choice is random search, where \citet{rijn-kdd18a} replace the uniform prior with a prior learned from hyperparameter optima on other datasets. An interesting line of work also aims to allow the users to manually specify their prior to guide BO~\citep{souza-ecml21a,anonymous-iclr22a}.

\paragraph{Scalable models for standard HPO}
Bayesian optimization has also been done with models that scale better with the number of observations, such as random forests~\citep{hutter-lion11a}, Parzen estimators~\citep{bergstra-nips11a} and neural networks~\citep{snoek-icml15a,springenberg-nips16a}. These models come with challenges of their own, such as poor uncertainty extrapolation with limited observations and hyperparameter sensitivity. There are also extensions of GPs for large datasets, most notably sparse GPs~\citep{csato-np02a}. However, sparse GPs can also have poor uncertainty extrapolation~\citep{bauer-neurips16a,wang-aistats18a} and were shown to perform less well for Bayesian optimization than standard GPs~\citep{mcintire-uai16a}. Recently, there has been interest in scaling GPs by only optimizing in subspaces~\citep{eriksson-neurips19a}, however, it is unclear how to combine such approaches with transfer HPO. Therefore, GPs remain the standard model for practical Bayesian optimization, especially since most methods to jointly model several tasks require the definition of meta-features or expensive calculation of task similarities. 

\section{Discussion and Conclusion}\label{sec:conclusion}

Our goal was to have a transfer hyperparameter optimization method that scales beyond standard GPs, is hyperparameter free, does not require meta-features, and guarantees to not perform substantially worse than standard BO. We showed that a linear combination of GPs, weighted by their posterior probability of describing the observations, and probabilistically dropping underperforming models, gives rise to several novel methods to fulfill these four desiderata. Our experiments on six hyperparameter optimization benchmarks showed that the proposed methods performed better than alternatives from the literature, and they are scalable and effective methods for conducting transfer learning in GP-based BO.
Because of the weight dilution prevention we also obtain a performance guarantee in terms of simple regret. In particular, using both our new \rgpe{} weighting scheme and our novel weight dilution prevention mechanism together with the transfer acquisition function leads to very stable performance, and we recommend this \tafrgpe{} combination as the new default algorithm for transfer HPO.

A side-effect of our study is the better understanding of the previous transfer HPO methods 
\tstr{}~\citep{wistuba-ecml16a} and TAF~\citep{wistuba-ml18a}, which we showed are special cases of the linear combination of Gaussian processes proposed in this work.

Furthermore, we have made available the largest collection of transfer HPO methods and benchmarks in a single framework to date with a total of six benchmarks and 9 competitor methods from the literature. This will allow future researchers to more easily conduct research on transfer HPO by having a set of benchmarking problems and baselines to compare against readily available.

Finally, we acknowledge limitations of the proposed method which give directions for future work. First, they are limited to predictive models and cannot be used with models like the \emph{Tree Parzen Estimator}~\citep{bergstra-nips11a}. Second, our evaluations are limited to spaces that can be modeled well by regular Gaussian processes, which does not include the complex spaces of, e.g., Auto-sklearn~\citep{feurer-nips2015a}, Auto-WEKA~\citep{thornton-kdd13a}, or Auto-PyTorch~\citep{zimmer-tpami21a}. As the only requirement for the model is being a probabilistic model, we expect that using random forests or neural networks for these is a promising direction. Third, all methods introduced hinge on the underlying BO library as they converge to vanilla BO. If the BO library cannot provide excellent performance by itself, putting transfer learning on top of it does not guarantee that it would outperform a well-tuned BO tool. Finally, it would be interesting to extend the proposed methods to other acquisition functions to explicitly handle noise~\citep{letham-ba19a} or work in a multi-fidelity setting~\citep{wu-uai20a}.

\subsubsection*{Acknowledgements}
Thanks to Till Varoquaux for support in developing the method, Sam Daulton for help with the implementation, Martin Wistuba for help with the \tstr{} and TAF baselines, and Rodolphe Jenatton and Valerio Perrone for help with the \ablr{} baseline.  We thank Josif Grabocka and the anonymous reviewers of previous versions for helpful feedback on the manuscript, and Carl Hvafner and Luigi Nardi for feedback on the theory. The authors acknowledge support by the state of Baden-Württemberg through bwHPC and the German Research Foundation (DFG) through grant no INST 39/963-1 FUGG (bwForCluster NEMO).

\bibliography{strings,lib,local,proc}

\newpage

\appendix

\section{Relation between TAF, SFMO and Active Testing}\label{app:active_testing}

\citet{leite-mldmpr12} introduced \emph{active testing} which uses relative landmarking (pairwise ranking of all observations so far) to decide which configuration $\mathbf{x}$, from a finite set of choices, to run next. For each candidate configuration $\mathbf{x}$ that was not yet run on the target dataset, they do a table lookup to determine whether and by how much that algorithm has improved over the current best observed algorithm on each meta-dataset, and weight its improvement by a task similarity measure computed using the pairwise ranking of all observations so far. More specifically, active testing selects the next algorithm according to
\begin{equation}\label{equ:active_testing}
    \argmax_{\mathbf{x}} \sum_{i=1}^{t-1} RL(\mathbf{x}^i, \mathbf{x}_{best}^i) Sim(t, i)
\end{equation}
with RL being a \emph{relative landmark} defined as
\begin{align}
    &RL(\mathbf{x}^i_k, \mathbf{x}_{best}^i) \nonumber \\
    &= \mathbbm{1} (y^i_k < y^i_{best})(y^i_{best} - y^i_k)) \nonumber  \\
    &= \max (0, y_{best}^i - y^i_k))) \nonumber  \\
    &= I(\mathbf{x}^i_k), \label{equ:active_testing:improvement}
\end{align}
$\mathbf{x}_{best}$ being the best algorithm observed on the target task $t$ and $Sim(t,\cdot)$ is a similarity function defined as the fraction ranking pairs after a Laplace correction. We briefly clarify that $\mathbf{x}^i$ means the configuration $\mathbf{x}$ on dataset $i$, similarly $\mathbf{x}^i_k$ is the configuration indexed by $k$ on dataset $i$ and $y_k^i$ is the observed performance value.

Replacing the similarity $Sim(\cdot,\cdot)$ in Equation \ref{equ:active_testing} with the similarity divided by the sum of all similarities to the target model, $w_i = \frac{Sim(t,i)}{\sum_{j=1}^{t-1}Sim(t,j)}$, and also replacing the relative landmark in Equation \ref{equ:active_testing} with the improvement from Equation \ref{equ:active_testing:improvement} shows us that \emph{active testing} is similar to the \emph{transfer acquisition function} (TAF) with the following differences:
\begin{enumerate}
    \item Active Testing works only on a finite set of algorithms $a$, while TAF can be applied to problems with continuous hyperparameters.
    \item TAF models function values with a surrogate model, while active testing relies only on previously conducted function evaluations.
    \item TAF includes a model and an expected improvement term for the target task as well.
    \item The weighting function is slightly different to TAF, as it only takes the proportion of correct pairwise comparisons into account, but performs Laplace smoothing to compute $Sim(\cdot,\cdot)$. Most importantly, active testing does not include a bandwidth factor which determines whether a base model should be dropped.
    \item Active Testing models the improvement over the current best known design point, while TAF leaves the choice of $\mathbf{x}_{best}$ to the individual base models.
\end{enumerate}

This finding has two implications: First, we can use the weighting scheme we developed in the main paper in this setting as well, and would benefit from all improvements over \taftstr{} described in the text. Second, we have shown how to extend active testing to work on a continuous design space\footnote{In order to make this extension of active testing work on a design space which only contains discrete choices one would simply have to use a different model, such as matrix factorization models~\citep{fusi-neurips18a,yang-kdd2019a,yang-kdd2020a}} and incorporate a model for the target tasks.

If we instead drop the similarity term from Equation~\ref{equ:active_testing} and replace $y_{best}^i$ by $\min_{k \in (1,2,\dots,n)} y_k^i$, which means we consider improving over the performance of everything evaluated so far on dataset $i$, we obtain sequential model-free optimization described in Section~\ref{app:related}.

\section{Scaling Study}\label{sec:methodology:scaling}

\begin{figure}
    \centering
    \includegraphics[width=\textwidth]{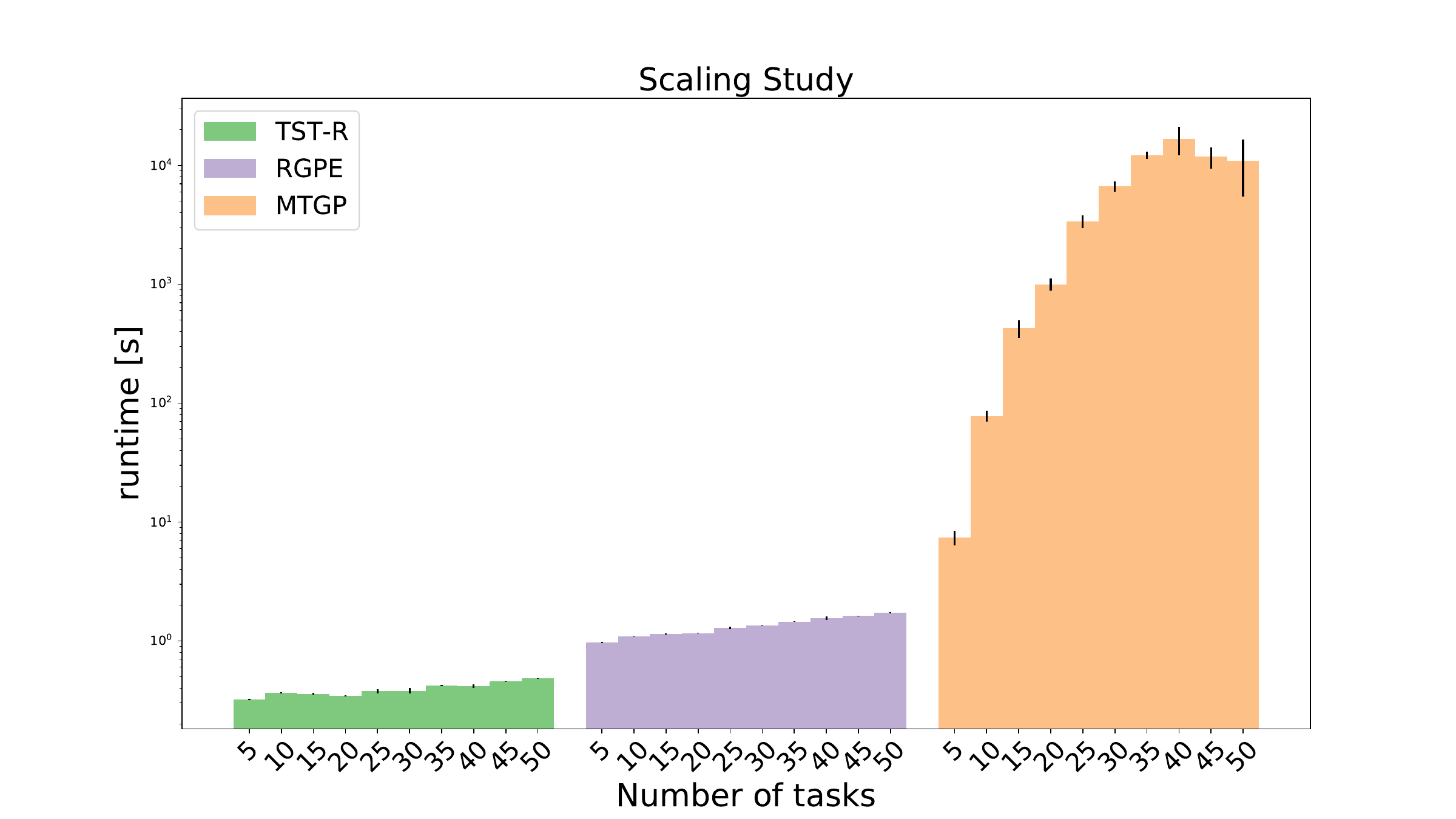}
    \caption{Scaling study comparing the proposed \rgpe{} to \tstr{} and multi-task GPs.}
    \label{fig:runtimes}
\end{figure}

In this section we compare the online training time of the \rgpe{} weighting scheme we proposed in Section 4 of the main paper with two competitor methods:
\begin{enumerate}
    \item \textbf{\tstr{}}~\citep[also see Section~\ref{sec:linear_combination_of_models}]{wistuba-ecml16a}, which for each model in the ensemble requires a prediction of the current observation and the computation of the loss function.
    \item \textbf{Multi-task Gaussian processes (MTGP)}~\citep[also see Section~\ref{app:related}]{swersky-nips13a}, which defines a covariance function $K((\mathbf{x}_k, i), (\mathbf{x}_l, j))$ between both the input location and the tasks, and put all observations on all tasks into a single Gaussian process. Task descriptions are then learned jointly with the kernel hyperparameters. The multi-task Gaussian process has a fitting complexity of $O(t^3n^3)$ and requires fitting and requires expensive hyperparameter tuning (see Appendix~\ref{app:related}). We use the implementation from BoTorch~\citep{balandat-neurips20a}, as it is, to the best of our knowledge, the only actively maintained implementation of a multi-task Gaussian process for Bayesian optimization.
\end{enumerate}
We give the fitting times on the \adaboost{} benchmark in Figure~\ref{fig:runtimes}. Each bar represents the average over five repetitions, and we used $5, 10, \dots, 50$ tasks for fitting. Furthermore, we used $50$ randomly sampled observations per task.

We can observe that \tstr{} and \rgpe{} are substantially faster than the multi-task Gaussian process, with \tstr{} always requiring less than a second to fit the whole ensemble. In contrast, \rgpe{} required at most 2 seconds to fit the whole ensemble for 50 tasks. In practice \rgpe{} requires less overhead as the fitting time is strongly influenced by the number of observations for the target task; we used 50 observations for the target task, while in practice we only observe these many observations at the very end of an optimization run. This overhead is a small premium one has to pay to obtain a hyperparameter-free transfer learning algorithm in Bayesian optimization.
Lastly, we can observe why using a single Gaussian process to fit all observations from all tasks does not scale to the size of benchmarks we consider here. The MTGP model requires more time to fit five tasks than the RPGE model for the highest number of tasks. Fitting the model takes roughly one hour for  25 tasks. The maximum time it required to fit the model was roughly 21000 seconds, which is almost six hours. Such an overhead rules out the usage of MTGP for the kind of problems we discuss in this paper. 

\section{Full Proof of Theorem 4.1}
\label{app:proof}

In this section, we show that the Bayesian optimization methods we introduced in the main paper which are built on our sampling-based weighting scheme are only a multiplicative factor slower than regular Bayesian optimization in the worst case. Specifically, we show that the convergence rates given by \citet{bull-jmlr11a} still hold and are slowed down by only a multiplicative factor. We will first restate the theoretical statement from the main paper. Second, we will discuss a statement for the specific case of functions with smoothness $\nu < 1$. Third, we will discuss a more general statement for functions with smoothness $\nu < \infty$.

\subsection{General Statement}

In each iteration, the proposed weighting mechanism has a positive chance of performing vanilla BO, in which only the target model is used. In these iterations, regular BO is conducted, and standard proofs for the convergence of BO therefore apply~\citep{bull-jmlr11a} with a slowdown factor.\footnote{We note that this only applies to results which provide guarantees in terms of the simple regret and not the cumulative regret.}

\noindent\textbf{Theorem 1} (From the main paper) \textit{
    Bayesian optimization using a linear combination of Gaussian processes with weights learned according to Section 4.2 of the main paper is at most a factor of 
    \begin{equation*}
        1 / \left( \frac{1}{H} \sum_{h=1}^{H} \left(\frac{h}{H}\right)^{t-1} \right)
    \end{equation*}
    slower than Bayesian optimization in the worst case.
}
As before, $H$ is the optimization horizon, while we change the number of observed data points $n^t$ to $h$, i.e. the current iteration.

\emph{Proof sketch:} In order for the proposed method to fall back to vanilla BO, we need the weights of all models except that of the base model to be zero. Given the definition of $p_{drop}$ in Equation \ref{equ:p_drop}, and setting $n_t$ to $h$ in $p_{drop}(i,h)$, we can calculate the probability of 
dropping all base models at step $h$ as $\prod_{i=0}^{t-1} p_{drop}(i,h)$,
and so the expected proportion of iterations that proceed as vanilla BO is

\begin{align}\
    & \frac{1}{H} \sum_{h=1}^{H} \prod_{i=1}^{t-1} p_{drop}(i,h)\\
    & = \frac{1}{H} \sum_{h=1}^{H} \prod_{i=1}^{t-1} \left(1 - \left(\left(1 - \frac{h}{H}\right) \frac{\sum_{s=1}^S \mathbbm{1}(l_{i,s}  < l_{t,s})}{S}\right)\right)\\
    & \ge \frac{1}{H} \sum_{h=1}^{H} \left(1 - \left(\left(1 - \frac{h}{H}\right) \frac{S}{S}\right)\right)^{t-1} \\
    & = \frac{1}{H} \sum_{h=1}^{H} \left(\frac{h}{H}\right)^{t-1}\\
    & > 0
\end{align}
The observations gathered in iterations when not all base models are dropped do not impose any issues on the convergence proof by \citet{bull-jmlr11a}. We will show this in the remainder of the section for fixed hyperparameters of a Gaussian process.

\subsection{\texorpdfstring{Functions with Smoothness $\nu < 1$}{Functions with Smoothness nu < 1}}

We briefly restate Theorem 2 from \cite{bull-jmlr11a}, which gives near-optimal convergence rates for GP-based Bayesian optimization with prior smoothness $\nu \leq 1$:

\noindent\textbf{Theorem 2} (From \citet{bull-jmlr11a}), \emph{Let $\pi$ be a prior with length-scale $\theta \in \mathcal{R}_+^d$. For any $R > 0$},
\begin{equation}
 L_n(EI(\pi), \mathcal{H}_\theta(X), R) = 
 \begin{cases}
  O(n^{-\nu / d} (\log n)^\alpha), & \nu \leq 1, \\
  O(n^{-1 / d}), & \nu > 1.
 \end{cases}
\end{equation}

The smoothness $\nu$ can be seen as a kernel (prior) hyperparameter to the Matérn kernel, and the kernel (prior) approaches the squared exponential covariance function for $\nu \rightarrow \infty$~\citep{rasmussen-book06a}. $L_n$ is the loss suffered over the ball $B_R$ in the reproducing Hilbert space $\mathcal{H}_\theta(X)$ after $n$ steps when applying an expected improvement strategy with given prior (kernel with hyperparameter values set to $\theta$):

\begin{align*}
    & L_n(EI(\pi),\mathcal{H}_\theta(X),R) \\ 
    & = \sup_{||f||_{H_\theta(X)}\leq R} \mathbb{E}^{EI(\pi)}_f[f(x^*_n) - \min f].
\end{align*}

In contrast to the main paper, we follow the notation from \citet{bull-jmlr11a} and use $f$ to denote the function we aim to minimize and $x$ to denote a single observation (in contrast to $\mathbf{x}$ in the main paper). We will describe symbols which are different from the notation in the main paper when introducing them. $\alpha$ is either zero for $\nu \notin \mathbb{N}$ or $0.5$ otherwise, 
and $R$ is an upper bound on the function norm, i.e. the maximum absolute value of the function $f$. $X$ is a subset of $\mathbb{R}^d$, and $x^*_n$ is the estimated optimizer of $f$ after $n$ steps.

\noindent\textbf{Proof of Bull's Theorem 2 still valid} The proof for Theorem 2 bounds the simple regret at some time step $n$ via several inequalities depending on Lemma 8 (which bounds the improvement over the current best observation based on a fitted GP), but the regret bound itself and its proof do not have a dependence on the Gaussian process conditioned on the observed data. Furthermore, the proof shows that there is a time step $n_k$, where the expected improvement is low and therefore BO must be close to the minimum. The existence of such a time step depends on Lemma 7, which states that there are at most $k$ time steps in which the posterior variance of the Gaussian process exceeds a certain value (which converges linearly to zero as we observe more data points). Together with an argument that BO can only improve $k$ times in total, Bull concludes that there must be a time step $k \leq k_n < 3k$ where the posterior variance at a future observation is small and some improvement occurs. We will therefore continue to discuss Lemma 7 and Lemma 8. We first start by restating Lemma 8:

\noindent\textbf{Lemma 8} (From \cite{bull-jmlr11a}) \emph{Let $||f||_{\mathcal{H}_\theta(X)} \leq R$. For $x \in X, n \in \mathbb{N}$, set $I = \max(0, f(x^*_n) - f(x))$, and $s = s_n(x, \theta)$. Then for}
\begin{equation*}
 \tau(x) := x \Phi(x) + \phi(x),
\end{equation*}
\emph{we have}
\begin{equation*}
 \max \left(I - Rs, \frac{\tau(-R/\sigma)}{\tau(R/\sigma)}I \right) \leq EI_n(x;\pi) \leq I + (R + \sigma)s,
\end{equation*}

where $s_n$ is the predictive standard deviation of the Gaussian process at iteration $n$ without the global scale of variation \citep[Equation 2]{bull-jmlr11a}, which is given by $\sigma$ in the above Lemma.

This bounds the expected improvement at a given point $x \in X$ for the case of the Gaussian process model predicting a variance larger than zero for $x$. The actual predictive distribution of the underlying Gaussian process changes with the observed data, but the Lemma is stated with respect to a Gaussian process conditioned on data, and therefore, Bull's proof holds for any sequence of previous observations $x_1,\dots,x_n$.

In contrast, Lemma 7 has a dependence on the sequence of data points observed.

\noindent\textbf{Lemma 7} (From \citet{bull-jmlr11a}) \emph{Set}
\begin{equation*}
 \beta := \begin{cases}
           \alpha, & \nu \leq 1, \\
           0, & \nu > 1.
          \end{cases}
\end{equation*}
\emph{Given $\theta \in \mathbb{R}_+^d$, there is a constant $C' > 0$ depending only on $X$, $K$ and $\theta$ which satisfies the following. For any $k \in \mathbb{N}$, and sequences $x_n \in X, \theta_n \geq \theta$, the inequality}
\begin{equation*}
 s_n(x_{n+1};\theta_n) \geq C'k^{-(\nu \wedge 1) / d}(\log k)^\beta
\end{equation*}
\emph{holds for at most $k$ distinct $n$.}

The proof of Lemma 7 consists of two parts. The first part shows that the posterior variance is bounded by the distance to the nearest point observed so far. As this describes the general behavior of a Gaussian process, it is a bound on the posterior variance given the observed data. The second part shows that most design points $x_{n+1}$ are close to a previous $x_i$, and that therefore the posterior variance of the Gaussian process model is effectively bounded. As we alter the strategy with which we choose the ``previous'' $x_i$, we need to ensure that this property still holds for points not chosen by the standard EI strategy. We fully restate this part of the proof:

We next show that most future evaluations $x_{n+1}$ are close to a previous $x_i$. $X$ is bounded, so can be covered by $k$ balls of radius $O(k^{-1/d})$. If $x_{n+1}$ lies in a ball containing some earlier point $x_i$, $i \leq n$, then we may conclude
\begin{equation*}
 s^2_n(x_{n+1};\theta_n) \leq C'^2k^{-2(\nu \wedge 1)/d}(\log k)^{2\beta},
\end{equation*}
for a constant $C' > 0$ depending only on $X$, $K$ and $\theta$. Hence as there are $k$ balls, at most $k$ points $x_{n+1}$ can satisfy
\begin{equation*}
 s_n(x_{n+1};\theta_n) \leq C'k^{-(\nu \wedge 1) / d} (\log k)^\beta.
\end{equation*}

Having stated this part of the original proof, we can see that the location of previously queried points is used to bound the predictive variance of the Gaussian process. However, whether a previously queried point in one of the $k$ balls was found by the initial design, EI with only the target GP, or BO with meta-learning does not matter for the posterior variance of the Gaussian process. 
We therefore conclude that BO methods based on the weighting mechanism introduced in the main paper take at most a multiplicative factor given by Theorem 4.1 longer to reach the time where most future evaluations $x_{n+1}$ are close to a previous $x_i$ compared to running only vanilla BO. This slowdown is exactly what we stated in Theorem 4.1.

As both Lemma 7 and Lemma 8 still hold, we conclude that Theorem 2 does not suffer from additional points observed between the iterations performing Bayesian optimization. As methods using the sampling-based weighting mechanism perform vanilla Bayesian optimization with probability $\frac{1}{H} \sum_{h=1}^{H} \left(\frac{h}{H}\right)^{t-1}$, these methods are at most a factor of
\begin{equation*}
 \frac{1}{\left( \frac{1}{H} \sum_{h=1}^{H} \left(\frac{h}{H}\right)^{t-1} \right)}
\end{equation*}
slower than vanilla Bayesian optimization.~~~~~~~~~~~~~~~~~~~~~~~~~~~~~~~~~~~~~~~~~~~~~~~~~~~~~~~~~~~~~~~~~~~~$\qedsymbol$

\subsection{\texorpdfstring{Functions with Smoothness $\nu < \infty$}{Functions with Smoothness nu < infinity}}

We briefly restate Theorem 5 from \citet{bull-jmlr11a}, which gives near-optimal convergence rates for GP-based Bayesian optimization in the more general case of prior smoothness $\nu < \infty$.

\noindent\textbf{Theorem 5} (Adapted from \citet{bull-jmlr11a} to only describe the case of a known prior $\pi$.), 
\emph{If $\nu < \infty$, then for any $R > 0$,}
\begin{equation*}
 L_n(EI(\pi,\epsilon),\mathcal{H}_{\theta^U}(X),R) = O((n / \log n)^{-\nu/d}(\log n)^\alpha),
\end{equation*}
\emph{while if $\nu = \infty$, the statement holds for all $\nu < \infty$}.

\noindent\textbf{Proof of Bull's Theorem 5 still valid}
Bull's proof of Theorem 5 works slightly differently than the proof of Theorem 2. Instead of relying on previously sampled points to bound the variance of the Gaussian process (and thereby also bound expected improvement and actual improvement), this proof relies on the observations being quasi-uniform in the search space. For this, the presented technique relies on falling back to a random sampling strategy with probability $0 < \epsilon < 1$.

The proof of Theorem 5 depends on Lemma 12, the proof for which preludes the proof of Theorem 5. It basically gives a probabilistic bound on the mesh norm of points chosen at random. More specifically, it gives the probability for $\sup_{x \in X} \min_{i=1}^n || x - x_i ||$ exceeding a certain value, which converges to zero with rate $\log n/n$. Conversely, the probability of having a low mesh norm increases with the number of randomly drawn samples, and having a small mesh norm is important for the remainder of Bull's proof. The proof of Lemma 12 does not take into account points selected by Bayesian optimization, but only points selected by random sampling, and therefore will not depend on points suggested by a meta-learning strategy.

Interestingly, the proof of Theorem 5 also does not depend on the points selected by Bayesian optimization, but rather states that posterior variances are small due to the quasi-uniform coverage of the search space, and therefore there exist times $n_k$ where the observed function values are close to $\min f$. The resulting bound is
\begin{equation*}
 \mathbb{E}_f^{EI(\pi,\epsilon)}[z^*_{2n+1} - z^*] \leq (2C'R + C'' (2R + \sigma))r_n.
\end{equation*}

As stated in Equation 2.1 and Sections 2.0 and 2.1 by \citet{bull-jmlr11a}, the left-hand side is the average-case performance at some future time $2n+1$, given by the expected loss under an expected improvement strategy with known hyperparameters of the Gaussian process. The quantities on the right-hand side are:
\begin{itemize}
 \item $r_n$: a constant depending on the current iteration $n$: $r_n = (n / \log n)^{-\nu / d}(\log n)^{\alpha}$.
 \item $R$: an upper bound on the function norm: $||f||_{\mathcal{H}_\theta(X)} \leq R$.
 \item $C'$: a constant $>0$, which multiplied with $r_n$ gives the upper bound on the probability of having chosen enough points at random, one point by expected improvement, and the mesh norm being small enough. \citet{bull-jmlr11a} states that this constant does not depend on $f$. As only a fraction of points are gathered by a combination of vanilla BO with every $\frac{1}{\epsilon}$th step sampled at random, it can take up to a factor given in Theorem 4.1 longer to observe the necessary points.
 \item $C''$: a constant $>0$, which multiplied with $r_n$ gives the upper bound on the predictive variance $s_n(x)$ of the Gaussian process for a given time step $n$, which in turn is bounded by the mesh norm. \citet{bull-jmlr11a} states that this constant depends only on $X$, $K$ and $C$ (a constant introduced in the proof of Lemma 12, which explains the random sampling behavior), therefore, it follows that it does not depend on $f$.
\end{itemize}

We conclude that Theorem 5 does not suffer from additional points observed in iterations where meta-learning is used and that the presented methods are at most a factor of 
\begin{equation*}
 \frac{1}{\left( \frac{1}{H} \sum_{h=1}^{H} \left(\frac{h}{H}\right)^{t-1} \right)}
\end{equation*}
slower than vanilla Bayesian optimization.~~~~~~~~~~~~~~~~~~~~~~~~~~~~~~~~~~~~~~~~~~~~~~~~~~~~~~~~~~~~~~~~~~~~$\qedsymbol$

\section{Experiments}\label{app:experiments}

In this section we give further details on the software used for this paper and details on the experiments to help reproducibility and replicability of our results. Furthermore, we give additional results on what scaling to use within the presented ensemble models, provide a sensitivity analysis of the hyperparameter of both the \tstrei{} and \taftstr{} method and finish with a comparison to our earlier work published in arXiv preprints.

\subsection{Search Spaces}

The search spaces for the surrogate benchmarks can be found in Tables \ref{tab:space-glmnet}, \ref{tab:space-svm} and \ref{tab:space-xgb} and the search space for the grid benchmarks can be found in Tables \ref{tab:space-adaboost_g}, \ref{tab:space-svm_g} and \ref{tab:space-nn}.

\begin{table}[ht!]
    \centering
    \small
\begin{tabular}{lcrr}
\toprule
Name & Range & Log & Cond \\
\midrule
$\alpha$ & $[0, 1]$ & No & No \\
$\lambda$ & $[2^{-10},2^{10}]$ & Yes & No \\
\bottomrule
\end{tabular}
\caption{Search space for the \glmnet{} surrogate benchmark.}\label{tab:space-glmnet}
\end{table}

\begin{table}[ht!]
    \centering
\begin{tabular}{lcrr}
\toprule
Name & Range & Log & Cond \\
\midrule
Kernel & \{Linear, Polynomial, Radial\} & No & No \\
$C$ & $[2^{-10}, 2^{10}]$ & Yes & Yes \\
$\gamma$ & $[2^{-10}, 2^{10}]$ & Yes & Yes \\
Degree & $[2, 5]$ & Yes & Yes \\
\bottomrule
\end{tabular}
\caption{Search space for the \svmfd{} surrogate benchmark.}\label{tab:space-svm}
\end{table}

\begin{table}[ht!]
    \centering
\begin{tabular}{lcrr}
\toprule
Name & Range & Log & Cond \\
\midrule
\#Rounds & $[1, 5000]$ & No & No \\
$\eta$ & $[2^{-10}, 2^0]$ & Yes & No \\
Subsample & $[0, 1]$ & No & No \\
Booster & \{GBLinear, GBTree\} & No & No \\
Max depth & $\{1, 15\}$ & No & Yes \\
Min child weight & $[2^0, 2^7]$ & Yes & Yes \\
Colsample by tree & $[0, 1]$ & No & Yes \\
Colsample by level & $[0, 1]$ & No & Yes \\
$\lambda$ & $[2^{-10}, 2^{10}]$ & Yes & No \\
$\alpha$ & $[2^{-10}, 2^{10}]$ & Yes & No \\
\bottomrule
\end{tabular}
\caption{Search space for the \xgb{} surrogate benchmark.}\label{tab:space-xgb}
\end{table}

\begin{table}[ht!]
    \centering
\begin{tabular}{lcrr}
\toprule
Name & Range & Log & Cond \\
\midrule
$I$ & $[2, 10000]$ & Yes & No \\
$M$ & $[2, 30]$ & Yes & No \\
\bottomrule
\end{tabular}
\caption{Search space for the \adaboost{} grid benchmark.}\label{tab:space-adaboost_g}
\end{table}

\begin{table}[ht!]
    \centering
\begin{tabular}{lcrr}
\toprule
Name & Range & Log & Cond \\
\midrule
Use linear kernel & \{No, Yes\} & No & No \\
Use polynomial kernel & \{No, Yes\} & No & No \\
Use RBF kernel & \{No, Yes\} & No & No \\
$C$ & $[2^{-5}, 2^6]$ & No & No \\
$d$ & $[2, 10]$ & No & No \\
$\gamma$ & $[0.0001, 1000]$ & Yes & No \\
\bottomrule
\end{tabular}
\caption{Search space for the \svmsd{} grid benchmark.}\label{tab:space-svm_g}
\end{table}

\begin{table}[ht!]
    \centering
\begin{tabular}{lcrr}
\toprule
Name & Range & Log & Cond \\
\midrule
Batch size & $[16, 256]$ & Yes & No \\
Learning rate & $[1e^{-4}, 1e^{-1}]$ & Yes & No \\
Momentum & $[0.1, 0.99]$ & No & No \\
Weight decay & $[1e^{-5}, 1e^{5}]$ & No & No \\
Number of layers & $[0.1, 0.99]$ & No & No \\
Maximum number of units per layer & $[16, 1024]$ & Yes & No \\
Dropout & $[0.0, 1.0]$ & No & No \\
\bottomrule
\end{tabular}
\caption{Search space for the Neural Network grid benchmark.}\label{tab:space-nn}
\end{table}

\subsection{Further Implementation Details}\label{ssec:implementation}

We now further discuss how we verified that our reimplementations of \tstr{} and TAF are correct.
The results reported in Figure 9 of \citet{wistuba-ml18a} were obtained by measuring the performance of Bayesian optimization with respect to multiple random seeds while keeping the random sample of configurations for training the base models the same across repetitions. This experimental design measures the performance with respect to one way the available meta-data is sampled. In contrast, we aim to measure the robustness of our method with respect to different ways the meta-data is obtained (i.e. every repetition having a different sample of configurations to train the base models on).
To ensure that our reimplementation is correct we strived to compare it to the original Java implementation in the same experimental setting. To allow a fair comparison we modified the implementation of \tstr{}~\cite{wistuba_code} to sample the configurations for each repetition based on the random seed. Then, we ran both our reimplementation and the original implementation with 25 different random seeds on the \svmsd{} benchmark and observed that both methods perform highly similar and conclude that our reimplementation is correct.

\subsection{Standardization, Copula Transform or Raw Values?}\label{ssec:copula}

As the results for different tasks can live on different scales, the different methods need to take this into account. We discussed in Section~\ref{sec:rgpe} that we standardize the observations $\{y_k^i\}_{k=1}^{n_i}$ separately per task $i \in (1, \dots, t)$ so that for each task the mean of the observations is zero and the variance is one. However, it has so far not been studied whether this is necessary or one can also use the raw values. Moreover, we are interested in whether the Copula transformation that was recently proposed as an alternative scaling for meta-learning in hyperparameter optimization is a generally applicable option~\citep{salinas-icml2020a}.

We apply the three different transformations to the proposed methods as well as to baseline methods: Our proposed \emph{\rgpenei{}} and \emph{\rgpemean{}} as well as the accompanying \emph{\tstr{}} baseline (including HPO); our proposed \emph{\rmogp{}} and \emph{\tafrgpe{}} as well as the \emph{\taftstr{}} baseline (including HPO); and also the \ablr{}~\citep{perrone-neurips18a} and \smfo{}~\citep{wistuba-icdm2015a,pfisterer-gecco21a} (see Section~\ref{app:related} for more information on these). We give these results in Table~\ref{tab:experiment:copula}.

Methods that combine on a model level (\emph{\rgpenei{}}, \emph{\rgpemean{}}, \emph{\tstr{}}, \emph{\ablr{}} and \emph{\smfo{}}) use the zero mean unit variance scaling as their default. In contrast, the methods that combine the individual models on the acquisition function level do not scale by default. Looking at the former, using no scaling on average results in worse performance than using the default scaling. However, we find that using the Copula transform can lead to further improvements, and except for two methods on the \adaboost{} benchmark, always performs best or not significantly worse than the best scaling for the respective method. Looking at the latter, not scaling the data at all appears to be a good default. The Copula appears to perform en par for the two TAF versions, and appears to be slightly better for \rmogp{}. Finally, the zero mean unit variance scaling performs rather well, too, but the other two methods have a slight edge.

Interestingly, the Copula transform appears to work very well on average, but leads to substantially worse results on the \svmfd{} benchmark and can lead to slightly worse results for the TAF methods on the \nn{} benchmark. We checked what caused these performance drops and found that for \svmfd{} this was caused by a single dataset. Consequently, we checked that the target model was picked as the sole model for sufficient amount of time, and found that this is the case. Therefore, we attribute this failure to our underlying GP which was not able to escape this local minimum (most likely due to the conditional nature of the hyperparameter space). For the \nn{} benchmark we cannot fully attribute this failure to a single dataset.

\begin{table}[htbp]
    \centering
    \small
    
\begin{tabular}{
l
p{0.11\textwidth}@{\hskip 0mm}
p{0.11\textwidth}@{\hskip 0mm}
p{0.11\textwidth}@{\hskip 0mm}
p{0.11\textwidth}@{\hskip 0mm}
p{0.11\textwidth}@{\hskip 0mm}
p{0.11\textwidth}
}
\toprule
{} & \rotatebox[origin=b]{50}{\glmnet{}} & \rotatebox[origin=b]{50}{\svmfd{}} & \rotatebox[origin=b]{50}{\nn{}} & \rotatebox[origin=b]{50}{\xgb{}} & \rotatebox[origin=b]{50}{\adaboost{}} & \rotatebox[origin=b]{50}{\svmsd{}} \\
\midrule
\rgpemeanvar{}*           & \textbf{2.20} & \textbf{0.37} & 2.71 & 0.64 & 0.62 & \textbf{0.39} \\
\rgpecopula{}   &  \underline{\textbf{0.13}} &              \textbf{1.41} &  \underline{\textbf{2.13}} &  \underline{\textbf{0.60}} & \textbf{0.57} &  \underline{\textbf{0.32}} \\
\rgpeunscaled{} & \textbf{2.70} &  \underline{\textbf{0.34}} &          3.05 &          0.94 &  \underline{\textbf{0.49}} & \textbf{0.34} \\
\midrule
\rgpemeanmeanvar{}*          &          0.23 &  \underline{\textbf{0.26}} &          2.55 &          0.65 & \textbf{0.55} & \textbf{0.47} \\
\rgpemeancopula{}   & \textbf{0.20} & \textbf{1.55} &  \underline{\textbf{2.22}} &  \underline{\textbf{0.62}} &          0.66 &  \underline{\textbf{0.42}} \\
\rgpemeanunscaled{} &  \underline{\textbf{0.15}} &          0.62 &          2.94 &          0.64 &  \underline{\textbf{0.44}} & \textbf{0.43} \\
\midrule
\tstrhpomeanvar{}* & \textbf{0.21} &  \underline{\textbf{0.27}} &          3.19 & \textbf{0.61} & \textbf{0.57} & \textbf{0.55} \\
\tstrhpocopula{} & \textbf{0.29} &          2.17 &  \underline{\textbf{2.69}} &  \underline{\textbf{0.57}} &  \underline{\textbf{0.49}} &  \underline{\textbf{0.53}} \\
\tstrhpounscaled{} &  \underline{\textbf{0.18}} & \textbf{0.60} &          4.73 &          0.64 & \textbf{0.64} & \textbf{0.59} \\
\midrule
\taftstrhpomeanvar{} & \textbf{0.10} & \textbf{0.58} &  \underline{\textbf{1.67}} &          0.66 &          0.68 & \textbf{0.49} \\
\taftstrhpocopula{} &  \underline{\textbf{0.08}} & \textbf{1.13} & \textbf{1.85} &  \underline{\textbf{0.53}} & \textbf{0.60} &  \underline{\textbf{0.47}} \\
\taftstrhpounscaled{}* & \textbf{0.17} &  \underline{\textbf{0.50}} & \textbf{1.71} &          0.62 &  \underline{\textbf{0.57}} &          0.50 \\
\midrule
\tafrgpemeanvar{} &  \underline{\textbf{0.14}} & \textbf{1.30} &          2.06 &  \underline{\textbf{0.57}} & \textbf{0.65} &  \underline{\textbf{0.35}} \\
\tafrgpecopula{}  &  \underline{\textbf{0.14}} & \textbf{1.30} &          2.06 &  \underline{\textbf{0.57}} & \textbf{0.65} &  \underline{\textbf{0.35}} \\
\tafrgpeunscaled{}*         & \textbf{0.16} &  \underline{\textbf{0.38}} &  \underline{\textbf{1.55}} &          0.67 &  \underline{\textbf{0.63}} & \textbf{0.45} \\
\midrule
\rmogpmeanvar{} & \textbf{0.17} & \textbf{0.38} & \textbf{1.99} &          0.71 &  \underline{\textbf{0.73}} & \textbf{0.53} \\
\rmogpcopula{}      & \textbf{0.16} & \textbf{1.28} &  \underline{\textbf{1.89}} &  \underline{\textbf{0.56}} & \textbf{0.76} &  \underline{\textbf{0.37}} \\
\rmogpunscaled{}* &  \underline{\textbf{0.16}} &  \underline{\textbf{0.30}} & \textbf{1.93} &          0.80 & \textbf{0.76} & \textbf{0.58} \\
\midrule
\ablrmeanvar{}*         &          2.95 &          0.86 &  \underline{\textbf{5.03}} &          1.58 &  \underline{\textbf{0.52}} & \textbf{1.86} \\
\ablrcopula{} &  \underline{\textbf{0.19}} &  \underline{\textbf{0.61}} & \textbf{5.18} &  \underline{\textbf{1.19}} &          0.72 &  \underline{\textbf{1.71}} \\
\midrule
\smfomeanvar{}*         &  \underline{\textbf{0.25}} & \textbf{1.74} &  \underline{\textbf{1.75}} &  \underline{\textbf{0.87}} & \textbf{1.11} & \textbf{1.21} \\
\smfocopula{} & \textbf{0.30} &  \underline{\textbf{1.67}} & \textbf{1.85} & \textbf{0.90} &  \underline{\textbf{0.71}} &  \underline{\textbf{1.08}} \\
\bottomrule
\end{tabular}
\caption{Results comparing standardization~\citep[mean/var]{yogatama-aistats14a}, the Copula transform~\citep[Copula]{salinas-icml2020a} and no transformation (unscaled) for \rgpenei{}, \rgpemean{}, \tstrei{}, \taftstr{}, \tafrgpe{}, \rmogp{}, \ablr{} and \smfo{}. We mark the method's default by an asterisk. The numbers reported are the average normalized regret~\citep{wistuba-ml18a}. We boldface the best value per benchmark and underline methods that are not significantly different according to a Wilcoxon signed-rank test with $\alpha=0.05$~\citep{demsar-06a}.}
    \label{tab:experiment:copula}
\end{table}

\subsection{Sensitivity Analysis for \tstrei{} and \taftstr{} Methods}\label{ssec:ablation_tstr_taf}

In this subsection we study certain aspects of the baseline methods \tstrei{} and \taftstr{}. Specifically, we are interested in effect of weight dilution prevention on both methods and in how to set the hyperparameter $\rho$.

We compare both methods with and without weight dilution prevention in Table~\ref{tab:ablation-tstr-taf-wd}.

\begin{table}[tb]
    \centering
    \small

\begin{tabular}{
l
p{0.11\textwidth}@{\hskip 0mm}
p{0.11\textwidth}@{\hskip 0mm}
p{0.11\textwidth}@{\hskip 0mm}
p{0.11\textwidth}@{\hskip 0mm}
p{0.11\textwidth}@{\hskip 0mm}
p{0.11\textwidth}
}
\toprule
{} & \rotatebox[origin=b]{50}{\glmnet{}} & \rotatebox[origin=b]{50}{\svmfd{}} & \rotatebox[origin=b]{50}{\nn{}} & \rotatebox[origin=b]{50}{\xgb{}} & \rotatebox[origin=b]{50}{\adaboost{}} & \rotatebox[origin=b]{50}{\svmsd{}} \\
\midrule
\tstrei{}     &              \textbf{0.20} &                       0.49 &  \underline{\textbf{3.00}} &              \textbf{0.78} &  \underline{\textbf{0.52}} &  \underline{\textbf{0.64}} \\
\tstrwd{} &  \underline{\textbf{0.14}} &  \underline{\textbf{0.43}} &              \textbf{3.01} &  \underline{\textbf{0.72}} &                       0.59 &              \textbf{0.76} \\
\midrule
\taftstr{}    &  \underline{\textbf{0.17}} &  \underline{\textbf{0.48}} &              \textbf{2.95} &  \underline{\textbf{0.64}} &  \underline{\textbf{0.73}} &              \textbf{0.46} \\
\taftstrwd{} &              \textbf{0.17} &              \textbf{0.60} &  \underline{\textbf{2.84}} &              \textbf{0.67} &              \textbf{0.89} &  \underline{\textbf{0.40}} \\
\bottomrule
\end{tabular}
\caption{Ablation study on the effect of weight dilution for \tstrei{} and \taftstr{}. The numbers reported are the average normalized regret~\citep{wistuba-ml18a}. We boldface the best value per benchmark and number of function evaluations and underline methods that are not significantly different according to a Wilcoxon signed-rank test with $\alpha=0.05$~\citep{demsar-06a}; separately for \tstrei{} and \taftstr{}}
    \label{tab:ablation-tstr-taf-wd}
\end{table}

\begin{table}[tb]
    \centering
    \small

\begin{tabular}{
l
p{0.11\textwidth}@{\hskip 0mm}
p{0.11\textwidth}@{\hskip 0mm}
p{0.11\textwidth}@{\hskip 0mm}
p{0.11\textwidth}@{\hskip 0mm}
p{0.11\textwidth}@{\hskip 0mm}
p{0.11\textwidth}
}
\toprule
{} & \rotatebox[origin=b]{50}{\glmnet{}} & \rotatebox[origin=b]{50}{\svmfd{}} & \rotatebox[origin=b]{50}{\nn{}} & \rotatebox[origin=b]{50}{\xgb{}} & \rotatebox[origin=b]{50}{\adaboost{}} & \rotatebox[origin=b]{50}{\svmsd{}} \\
\midrule
\tstrei{}                &              \textbf{0.20} &                       0.49 &              \textbf{3.00} &                       0.78 &              \textbf{0.52} &              \textbf{0.64} \\
\tstrhpo{}           &              \textbf{0.21} &              \textbf{0.27} &                       3.24 &              \textbf{0.61} &              \textbf{0.58} &                       0.55 \\
\tstrhpoglmnet{} &  \underline{\textbf{0.14}} &                       0.49 &  \underline{\textbf{2.97}} &              \textbf{0.77} &  \underline{\textbf{0.46}} &              \textbf{0.58} \\
\tstrhposvmfd{}    &              \textbf{0.20} &  \underline{\textbf{0.27}} &              \textbf{3.00} &                       0.78 &              \textbf{0.52} &              \textbf{0.64} \\
\tstrhpoxgb{}    &              \textbf{0.15} &                       0.51 &                       3.21 &  \underline{\textbf{0.61}} &              \textbf{0.51} &                       0.63 \\
\tstrhpoadaboost{}      &  \underline{\textbf{0.14}} &                       0.49 &  \underline{\textbf{2.97}} &              \textbf{0.77} &  \underline{\textbf{0.46}} &              \textbf{0.58} \\
\tstrhposvmsd{}           &              \textbf{0.17} &              \textbf{1.06} &                       3.57 &                       0.81 &                       0.61 &  \underline{\textbf{0.51}} \\
\tstrhponn{}            &  \underline{\textbf{0.14}} &                       0.49 &  \underline{\textbf{2.97}} &              \textbf{0.77} &  \underline{\textbf{0.46}} &              \textbf{0.58} \\
\midrule
\taftstr{}               &                       0.17 &                       0.48 &                       2.95 &              \textbf{0.64} &              \textbf{0.73} &              \textbf{0.46} \\
\taftstrhpo{}           &                       0.17 &              \textbf{0.50} &                       1.72 &              \textbf{0.62} &              \textbf{0.56} &              \textbf{0.49} \\
\taftstrhpoglmnet{} &  \underline{\textbf{0.12}} &                       0.50 &              \textbf{1.73} &                       0.68 &              \textbf{0.71} &              \textbf{0.44} \\
\taftstrhposvmfd{}    &              \textbf{0.16} &  \underline{\textbf{0.37}} &                       3.08 &                       0.87 &              \textbf{0.51} &              \textbf{0.53} \\
\taftstrhpoxgb{}    &              \textbf{0.18} &                       0.53 &              \textbf{1.63} &  \underline{\textbf{0.62}} &              \textbf{0.78} &              \textbf{0.40} \\
\taftstrhpoadaboost      &              \textbf{0.16} &              \textbf{0.45} &                       2.84 &              \textbf{0.67} &  \underline{\textbf{0.44}} &              \textbf{0.50} \\
\taftstrhposvmsd{}           &              \textbf{0.18} &              \textbf{0.43} &              \textbf{1.63} &              \textbf{0.67} &              \textbf{0.89} &  \underline{\textbf{0.34}} \\
\taftstrhponn{}            &                       0.16 &                       0.60 &  \underline{\textbf{1.58}} &              \textbf{0.64} &              \textbf{0.75} &              \textbf{0.40} \\
\bottomrule
\end{tabular}

\caption{Ablation study for how to set the hyperparameter $\rho$ of \tstrei{} and \taftstr{}. The numbers reported are the average normalized regret~\citep{wistuba-ml18a}. We boldface the best value per benchmark and number of function evaluations and underline methods that are not significantly different according to a Wilcoxon signed-rank test with $\alpha=0.05$~\citep{demsar-06a}; separately for \tstrei{} and \taftstr{}}
    \label{tab:ablation-tstr-taf-hpo}

\end{table}

Lastly, we study the effect of tuning the hyperparameter $\rho$ for \tstrei{} and \taftstr{} and show results in Table~\ref{tab:ablation-tstr-taf-hpo}. In total, we compare three different strategies to set $\rho$:
\begin{itemize}
    \item[\tstrei{}:] The hyperparameter value that has the best average result on the remaining five benchmarks.
    \item[\tstrhpo{}:] Leave-one-task-out hyperparameter tuning. For this, we conducted a grid search on the all but $t-1$ tasks of a benchmark and applied the best found hyperparameter setting to the target task.
    \item[\tstr{}(EI,benchmark name):] The hyperparameter value that is best on all tasks of the given benchmark. This should be overly optimistic when applied to the same benchmark.
\end{itemize}

For \tstrei{}  we observe that tuning the hyperparameters using the leave-one-task-out procedure directly on the benchmark does not necessarily give an advantage over using a fixed setting. Tuning directly on the remaining tasks of the benchmark improves in only three out of six benchmarks (\svmfd{}, \xgb{} and \svmsd{}) while it results in roughly the same performance for another two (\glmnet{} and \adaboost{}) and a performance degradation for the \nn{} benchmark. Interestingly, it never arrives at a value as good as choosing the hyperparameter on all $t$ tasks of a benchmark, and in two cases it is significantly worse.
When looking at the performance of transferring the hyperparameter setting from a single benchmark, we find quite different results for the different benchmarks. For \glmnet{} the transfer works from all other benchmarks, while for \svmfd{} only the transfer from \svmsd{} gives significantly similar results as the best setting. For the \nn{} benchmark three out of five settings give good results, while two give results that are significantly worse than the best.

Conversely, we can observe a strong influence of hyperparameter optimization on  \taftstr{}. Tuning hyperparameters using the leave-one-task-out procedure directly on the benchmark improves two out of six benchmarks and ties on the remaining six, and tuning hyperparameters directly on the benchmark is not significantly worse picking the hyperparameter value on the remaining tasks. The most drastic change is for the neural network benchmark, where \taftstr{} improves substantially and becomes very competitive with the proposed \rgpe{}-based methods (see Tables~\ref{tab:results-glm-svm}, \ref{tab:results-nn-xgb} and \ref{tab:results-ada-svm}). Furthermore, we find that for \taftstr{} all methods of choosing $\rho$ work for the grid-based benchmarks \adaboost{} and \svmsd{}, while success is not guaranteed for the remaining four benchmarks. 

Based on these results, we find that \taftstr{} can be a competitive method, but is very sensitive to the setting of its hyperparameters $\rho$. If tuned correctly it drastically outperforms its competitor \tstrei{} in terms of raw performance.

We would like to highlight that none of these hyperparameter optimization would be practically feasible, as they require tremendous amount of computation. For the hyperparameter value that was best on all other benchmarks we required access to five auxiliary benchmarks and needed to do meta-level hyperparameter optimization on them. For the leave-one-task-out hyperparameter tuning we would only require access to the remaining $t - 1$ tasks and run Bayesian optimization with all meta-level hyperparameters on it. The hyperparameter that is tuned based on a single benchmark has similar costs in that it requires to solve $t$ tasks for each hyperparameter value. While the overhead of the BO method is negligible, real function calls to the algorithm to be optimized make such procedure extremely expensive. We were only able to conduct this hyperparameter optimization of the hyperparameter optimization algorithm because the calls to the target algorithms were either pre-recorded or simulated by a surrogate benchmark. In some settings it might also not be possible to access the base tasks any more, for example due to data retention rules.

\subsection{Comparison to Earlier Versions}\label{app:older}

For the sake of completeness, in this section we compare against previous implementations of the methods in this paper that were laid out in earlier, publicly available preprints. Concretely, we compare the \rgpe{} version from this paper (\rgpenei{}) against the closed-form noisy EI~\citep[\cfnei{}]{feurer-arxiv2021a} and noisy EI, using posterior samples from the Gaussian processes to compute the weightings and a 95-quantile weight-dilution strategy~\citep[\rgpequantile{}]{feurer-arxiv2018a}, and give results in Table~\ref{tab:experiments:old}.\footnote{We would like to highlight that Equation 2 of \cite{feurer-arxiv2018a} was used with a noise-free Gaussian process. In the more general case of also tuning the noise hyperparameter of the Gaussian process that we consider in this work we employ Equation~\ref{eq:loss_target} from this paper.} We can observe that the latest implementation of our method is either the best or not significantly worse than the best on all six benchmark problems. The older competitors are not significantly worse, either, except on the neural network benchmark. Nevertheless, since the \rgpenei{} from this paper is substantially worse than \tafrgpe{} we also introduced in this work, we conclude that we have made significant improvements over these earlier versions.

\begin{table}[htbp]
    \centering
    \small
\begin{tabular}{l@{\hskip 2mm}cccccc}

\toprule
{} & \glmnet{} & \svmfd{} & \nn{} & \xgb{} & \adaboost{} & \svmsd{} \\
\midrule
\cfnei{} &  \underline{\textbf{0.16}} &              \textbf{0.54} &                       2.94 &              \textbf{0.68} &              \textbf{0.67} &              \textbf{0.44} \\
\rgpequantile{} &              \textbf{0.21} &              \textbf{0.45} &                       3.28 &              \textbf{0.78} &              \textbf{0.90} &              \textbf{0.41} \\
\rgpenei{}   &              \textbf{2.20} &  \underline{\textbf{0.37}} &  \underline{\textbf{2.71}} &  \underline{\textbf{0.64}} &  \underline{\textbf{0.62}} &  \underline{\textbf{0.39}} \\
\bottomrule
\end{tabular}

\caption{Comparison of the method proposed in this paper (\rgpenei{}) to previous versions proposed in earlier preprints~\citep{feurer-arxiv2018a,feurer-arxiv2021a}. The numbers reported are the average normalized regret~\citep{wistuba-ml18a}. We boldface the best value per benchmark and number of function evaluations and underline methods that are not significantly different according to a Wilcoxon signed-rank test with $\alpha=0.05$~\citep{demsar-06a}.}
    \label{tab:experiments:old}

\end{table}

\end{document}

%% file: macros.tex
\newcommand{\adaboost}{AdaBoost$_{g,2D}$}
\newcommand{\glmnet}{GLMNET$_{s,2D}$}
\newcommand{\svmsd}{SVM$_{g,6D}$}
\newcommand{\svmfd}{SVM$_{s,4D}$}
\newcommand{\xgb}{XGBoost$_{s,10D}$}
\newcommand{\nn}{NN$_{t,7D}$}

\newcommand{\numcompetitors}{$9$}

\newcommand{\pruningwithrandomsearch}{Random+Box}
\newcommand{\pruningwithBO}{GP(Box)}
\newcommand{\smfo}{SMFO}
\newcommand{\smfomeanvar}{SMFO(mean/var)}
\newcommand{\smfocopula}{SMFO(Copula)}
\newcommand{\wac}{WAC}
\newcommand{\klweighting}{KL-weighting}
\newcommand{\ablr}{ABLR}
\newcommand{\ablrmeanvar}{\ablr(mean/var)}
\newcommand{\ablrcopula}{\ablr(Copula)}
\newcommand{\gcpplusprior}{GCP+Prior}
\newcommand{\tstr}{TST-R}
\newcommand{\tstrei}{\tstr{}(EI)}
\newcommand{\tstrwd}{\tstr(EI,wd)}
\newcommand{\tstrhpomeanvar}{\tstr{}(EI,HPO,mean/var)}
\newcommand{\tstrhpocopula}{\tstr{}(EI,HPO,Copula)}
\newcommand{\tstrhpounscaled}{\tstr{}(EI,HPO,unscaled)}
\newcommand{\tstrhpo}{\tstr{}(EI,HPO)}
\newcommand{\tstrhpoadaboost}{\tstr{}(EI,HPO,\adaboost)}
\newcommand{\tstrhpoglmnet}{\tstr{}(EI,HPO,\glmnet)}
\newcommand{\tstrhposvmsd}{\tstr{}(EI,HPO,\svmsd)}
\newcommand{\tstrhposvmfd}{\tstr{}(EI,HPO,\svmfd)}
\newcommand{\tstrhpoxgb}{\tstr{}(EI,HPO,\xgb)}
\newcommand{\tstrhponn}{\tstr{}(EI,HPO,\nn)}
\newcommand{\taftstr}{\tstr(TAF)}
\newcommand{\taftstrhpo}{\tstr(TAF,HPO)}
\newcommand{\taftstrhpoadaboost}{\tstr(TAF,HPO,\adaboost)}
\newcommand{\taftstrhpoglmnet}{\tstr(TAF,HPO,\glmnet)}
\newcommand{\taftstrhposvmsd}{\tstr(TAF,HPO,\svmsd)}
\newcommand{\taftstrhposvmfd}{\tstr(TAF,HPO,\svmfd)}
\newcommand{\taftstrhpoxgb}{\tstr(TAF,HPO,\xgb)}
\newcommand{\taftstrhponn}{\tstr(TAF,HPO,\nn)}
\newcommand{\taftstrwd}{\tstr(TAF,wd)}
\newcommand{\taftstrhpomeanvar}{\tstr{}(TAF,HPO,mean/var)}
\newcommand{\taftstrhpocopula}{\tstr{}(TAF,HPO,Copula)}
\newcommand{\taftstrhpounscaled}{\tstr{}(TAF,HPO,unscaled)}
\newcommand{\rgpe}{RGPE}
\newcommand{\rgpenei}{\rgpe(NEI)}
\newcommand{\rgpemean}{\rgpe(mean)}
\newcommand{\rgpeei}{\rgpe(EI)}
\newcommand{\cfnei}{\rgpe(CFNEI)}
\newcommand{\rgpequantile}{\rgpe(NEI,95)}
\newcommand{\rgpenowd}{\rgpe(NEI,None)}
\newcommand{\rgpemeanvar}{\rgpe(NEI,mean/var)}
\newcommand{\rgpecopula}{\rgpe(NEI,Copula)}
\newcommand{\rgpeunscaled}{\rgpe(NEI,unscaled)}
\newcommand{\rgpemeanmeanvar}{\rgpe(mean,mean/var)}
\newcommand{\rgpemeancopula}{\rgpe(mean,Copula)}
\newcommand{\rgpemeanunscaled}{\rgpe(mean,unscaled)}
\newcommand{\rmogp}{\rgpe(MoGP)}
\newcommand{\rmogpnowd}{\rgpe(MoGP,None)}
\newcommand{\rmogpmeanvar}{\rgpe(MoGP,mean/var)}
\newcommand{\rmogpcopula}{\rgpe(MoGP,Copula)}
\newcommand{\rmogpunscaled}{\rgpe(MoGP,unscaled)}
\newcommand{\rmogponehundred}{\rgpe{}(MoGP,$S=100$)}
\newcommand{\rmogponethousand}{\rgpe{}(MoGP,$S=1000$, default)}
\newcommand{\rmogptenthousand}{\rgpe{}(MoGP,$S=10000$)}
\newcommand{\rmogponehundredthousand}{\rgpe{}(MoGP,$S=100000$)}
\newcommand{\tafrgpe}{\rgpe(TAF)}
\newcommand{\tafrgpemeanvar}{\rgpe(TAF,mean/var)}
\newcommand{\tafrgpecopula}{\rgpe(TAF,Copula)}
\newcommand{\tafrgpeunscaled}{\rgpe(TAF,unscaled)}
\newcommand{\tafrgpenowd}{\rgpe(TAF,None)}